\documentclass{article} 
\usepackage{iclr2027_conference,times}

\usepackage{amsmath,amsfonts,bm}

\def\eqref#1{equation~\ref{#1}}

\def\1{\bm{1}}

\DeclareMathAlphabet{\mathsfit}{\encodingdefault}{\sfdefault}{m}{sl}
\SetMathAlphabet{\mathsfit}{bold}{\encodingdefault}{\sfdefault}{bx}{n}

\usepackage{graphicx}
\usepackage{algorithm}
\usepackage{algpseudocode}
\usepackage{url}
\usepackage{booktabs}
\usepackage{array}
\usepackage{longtable}
\usepackage{amsmath,amssymb}
\usepackage{etoolbox}
\usepackage{hyperref}
\hypersetup{hidelinks}

\renewcommand{\eqref}[1]{\textup{(\ref{#1})}}

\makeatletter
\patchcmd{\@maketitle}
  {\lhead{Published as a conference paper at ICLR 2027}}
  {\lhead{Preprint}}
  {}
  {\PackageError{SPARC}{Failed to patch ICLR header}{}}
\makeatother

\title{Shared Phase and Retention Control\\
for Efficient Adaptive Spectral Recurrence}

\author{
Wentao Wang$^{1}$ \quad
Hengyu Zhong$^{2}$ \quad
Yunhan Jiang$^{1}$ \quad
Jialiang An$^{1}$ \quad
Meng Lu$^{1}$\thanks{Corresponding author.}
\\
$^{1}$Peking University \quad $^{2}$GigaAI
}

\newcommand{\tabmean}[2]{%
  \ensuremath{\vcenter{\kern2pt\hbox{%
    \shortstack[c]{\ensuremath{#1}\\[-1pt]%
      {\normalfont\unboldmath\fontsize{7}{8}\selectfont$\pm\,#2$}}%
  }\kern2pt}}%
}

\newcommand{\tabnote}[1]{%
  \par\vspace{4pt}%
  {\footnotesize\raggedright #1\par}%
}

\newcommand{\tabstyle}{%
  \setlength{\tabcolsep}{4pt}%
  \renewcommand{\arraystretch}{1.22}%
  \setlength{\belowcaptionskip}{6pt}%
}

\newcommand{\tabgroup}{%
  \specialrule{0.3pt}{1.5pt}{1.5pt}%
}

\newcolumntype{C}[1]{>{\centering\arraybackslash}p{#1}}
\newcolumntype{L}[1]{>{\raggedright\arraybackslash}m{#1}}

\iclrfinalcopy

\begin{document}
\linespread{1.02}\selectfont

\maketitle

\begin{abstract}
Fixed-state spectral recurrences offer constant-memory inference and strong long-range retention through heterogeneous complex modes, but their static transitions cannot selectively filter distractors or adapt to evolving dynamics. Existing input-dependent recurrences introduce selectivity, yet real-valued gates lack adaptive rotation, while fine-grained complex transitions incur substantial projection and data-movement overhead and can expose persistent modes to cumulative drift. To address these limitations, we introduce Shared Phase and Retention Control for Efficient Adaptive Spectral Recurrence (SPARC), which modulates a layer's heterogeneous complex memory through two input-dependent scalars. Each mode retains its learned baseline timescale, frequency, and write response, while a retention-coupled clock damps phase corrections in persistent modes. The same shared signals regulate incoming content through modal gates and retention-dependent normalization. The resulting diagonal affine recurrence supports parallel associative scans for full-sequence backpropagation through time (BPTT) and exact structured Real-Time Recurrent Learning (RTRL), with sensitivity storage and updates linear in the recurrent layer's parameter count. We evaluate SPARC on partially observable continuous control, POPGym memory benchmarks, sequence classification, and language modeling. SPARC places among the top two means across all ten reinforcement-learning tasks. Compared with RG-LRU, it improves permuted-MNIST and sequential-CIFAR-10 accuracy by $10.13$ and $10.66$ percentage points, respectively, reduces held-out language-model perplexity by $2.00\%$, and accelerates mixer forward--backward computation by up to $1.69\times$ at width $D=2048$.
\end{abstract}


\section{Introduction}
\label{sec:introduction}

Sequence models across language, signal classification, and partially observable control must summarize long histories under tight compute and memory budgets. Transformers make past tokens accessible through content-dependent attention and support highly parallel training \citep{vaswani2017attention}, but dense global attention requires quadratic computation in sequence length and a linearly growing key--value cache during autoregressive inference \citep{de2024griffin,botev2024recurrentgemma}. Fixed-state recurrent models instead compress history into a continuously updated state, giving constant-memory inference and linear-time stream processing \citep{gu2020hippo}. Structured state-space models such as S4 \citep{gu2022s4}, the parallel scans of S5 \citep{smith2023s5}, and LRU's stable diagonal complex parameterization \citep{orvieto2023lru} make this approach practical for long sequences. RTU further extends spectral recurrence to online credit assignment through structured RTRL sensitivities, allowing memory to continue learning during interaction with a partially observable environment \citep{williams1989rtrl,elelimy2024rtu}.

Yet fixed spectral transitions evolve every step alike, whereas recurrent memory must decide which new observations to write and how stored information should evolve. Input-dependent gates provide selective retention \citep{gers2000forget,cho2014gru}, as in the linear time-varying dynamics of RG-LRU and Mamba \citep{de2024griffin,gu2023mamba}, while complex transitions in GateLoop and Mamba-3 additionally adapt a stored representation's rotation \citep{katsch2023gateloop,lahoti2026mamba3}. Their control pathways differ in how finely, and at what computational cost, they coordinate these updates across channels, heads, and state coordinates.

\begin{figure}[t]
    \centering
    \includegraphics[width=0.98\linewidth]{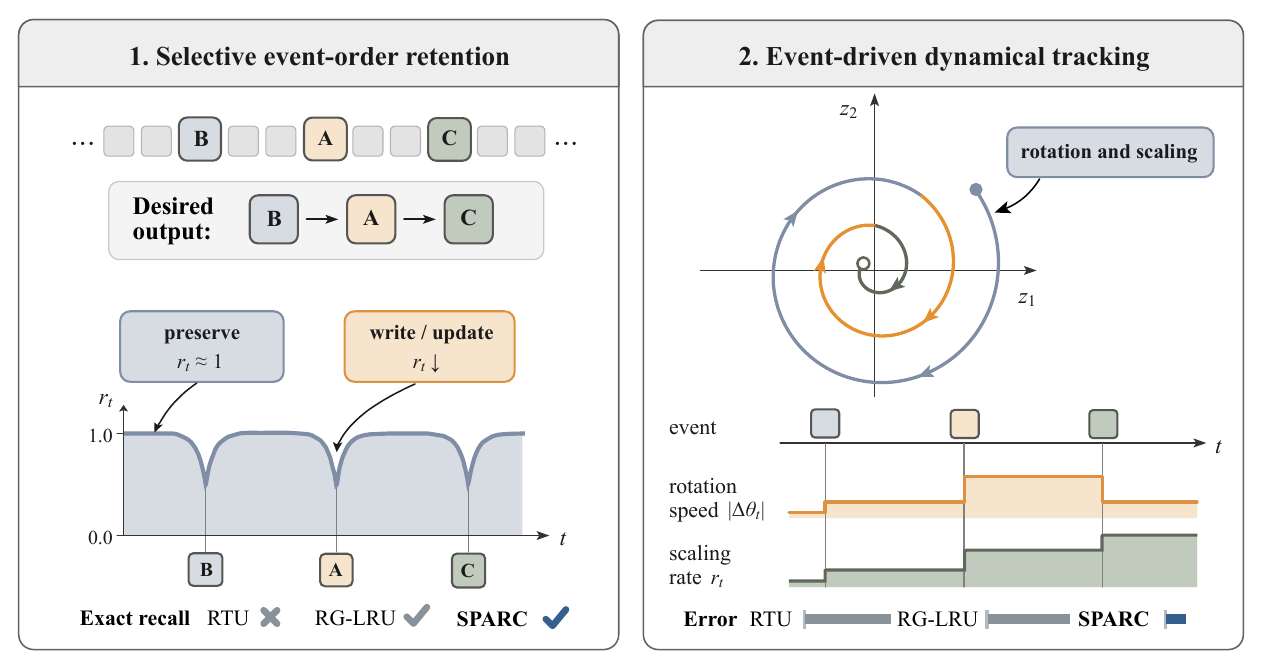}
    \caption{\textbf{Two complementary memory requirements.}
    Left: selective retention preserves event order among distractors; checks denote error-free recall. Right: event-driven contraction and rotation adapt stored-state dynamics; shorter bars denote lower tracking error. Appendix~\ref{app:controlled_diagnostics} gives the tasks and quantitative results.}
    \label{fig:motivation}
\end{figure}

The controlled tasks in Figure~\ref{fig:motivation} separate selective recall from adaptive state evolution. Retention and writing gates help preserve relevant event identities through distractors: RG-LRU and SPARC recover event order without error, whereas RTU has $16.99\%$ error. Tracking a latent system adds event-driven changes in contraction and rotation, requiring stored states to adapt both magnitude and phase. Here SPARC reaches NMSE $0.085$, compared with $0.345$ for RG-LRU, motivating adaptive phase control alongside selective retention.

Predicting retention and phase independently per mode increases control overhead and can disturb persistent memory. Sharing and low-rank projections reduce control costs in selective architectures \citep{gu2023mamba,dao2024mamba2,beck2024xlstm,peng2024eagle,yang2023gla}. A heterogeneous spectrum offers a complementary separation: distinct baseline timescales and frequencies supply memory diversity, while a smaller control space coordinates their input-dependent responses.

We introduce \emph{SPARC: Shared Phase and Retention Control for Efficient Adaptive Spectral Recurrence}, in which two layer-wide scalar signals separately modulate retention and phase across complex memory modes. Each mode retains its own decay, frequency, input projection, and write response. A retention-coupled phase clock attenuates rotation corrections in persistent modes, and the shared signals also regulate incoming content through modal writing gates and retention-dependent normalization.

Because transitions and writes depend only on the current input, the recurrence remains diagonal and affine in the previous state, supporting associative scans for full-sequence BPTT \citep{martin2018parallel,smith2023s5} and exact structured RTRL for recurrent-layer sensitivities \citep{williams1989rtrl,elelimy2024rtu}.

Our main contributions are:
\begin{enumerate}

    \item \textbf{Shared retention--phase control.}
    Two layer-wide scalars adapt retention and rotation while preserving modal timescales, frequencies, and write responses; a retention-coupled clock bounds phase drift in persistent modes while allowing transient modes to adapt rapidly.

    \item \textbf{Exact structured RTRL and efficient GPU kernels.}
    Exact sensitivities use $O(HD+H)$ work and trace storage per step. Fused writing and chunked scans reconstruct transitions from two control streams, yielding a $5.41\times$ speedup and $37.03\%$ lower memory than unshared control at $D=2048$, and up to $1.69\times$ speedup over GPU-optimized RG-LRU.

    \item \textbf{Advantages across tasks and training regimes.}
    SPARC leads mean returns on Walker-P, Hopper-P, Autoencode, and RepeatPrevious, reaching $0.817$ on RepeatFirst. Under BPTT, it improves pixel-sequence accuracy over RG-LRU by up to $10.66$ percentage points, exceeds spectral and selective baselines on Long Range Arena, and lowers language-model perplexity by $2.00\%$ while improving most zero-shot tasks.

\end{enumerate}


\section{Related Work}
\label{sec:related_work}

\subsection{Spectral Recurrence and Parallel Computation}
\label{sec:related_linear}
\label{sec:related_parallel}

S4 made long-range recurrence practical through structured state matrices \citep{gu2022s4}, after which S5 enabled parallel scans via a multi-input, multi-output formulation \citep{smith2023s5} and LRU simplified spectral modeling to stable diagonal complex dynamics \citep{orvieto2023lru}. Related spectral structures also underlie oscillatory dynamics in LinOSS \citep{rusch2025linoss} and online reinforcement learning in RTU \citep{elelimy2024rtu}. Although affine recurrences admit parallel prefix evaluation \citep{blelloch1990prefix,martin2018parallel}, hardware throughput depends heavily on coefficient generation and data movement: Mamba combines selective scans with kernel fusion and recomputation \citep{gu2023mamba}, while Mamba-2 organizes computation through state space duality \citep{dao2024mamba2}. SPARC retains parallel affine scans while keeping the dynamic control stream two-dimensional, reconstructing modal transitions within the scan.

\subsection{Input-Dependent Recurrent Dynamics}
\label{sec:related_dynamic}

Selective recurrences differ in how they parameterize retention and rotation across state coordinates. For real-valued retention, RG-LRU applies coordinate-wise gates \citep{de2024griffin}, RWKV-6 and GLA generate vector-valued decays through low-rank projections \citep{peng2024eagle,yang2023gla}, and Mamba-2 and mLSTM share scalar decays within each head \citep{dao2024mamba2,beck2024xlstm}. Mamba's input-dependent step size rescales heterogeneous coordinate rates within a channel and, for complex states, couples decay with rotation \citep{gu2023mamba}. Separate phase adaptation instead appears in GateLoop's directly predicted complex transitions \citep{katsch2023gateloop} and in Mamba-3's data-dependent rotations, which admit a rotary-embedding interpretation \citep{lahoti2026mamba3}. SPARC pairs a learned heterogeneous spectrum with two layer-wide retention and phase signals and a retention-coupled clock, separating control dimensionality from state capacity.

\subsection{Online Learning for Recurrent Models}
\label{sec:related_online}

Truncated BPTT bounds training memory by limiting the temporal credit-assignment window \citep{werbos1990bptt}. RTRL instead propagates sensitivities forward online, but dense state dependencies make exact updates expensive \citep{williams1989rtrl}. Approximate methods reduce this overhead through stochastic low-rank representations in UORO and KF-RTRL \citep{tallec2018uoro,mujika2018kfrtrl}, influence sparsity in SnAp \citep{menick2020sparse}, or local traces paired with learning signals in e-prop \citep{bellec2020eprop}. Structured recurrent layers take a complementary route, restricting state dependencies so that exact sensitivities remain compact \citep{zucchet2023online,elelimy2024rtu}. SPARC preserves this exact structure under adaptive retention and phase control through a diagonal state Jacobian, mode-local parameters, and shared controllers.


\section{Method}
\label{sec:method}

\begin{figure}[t]
    \centering
    \includegraphics[width=\linewidth]{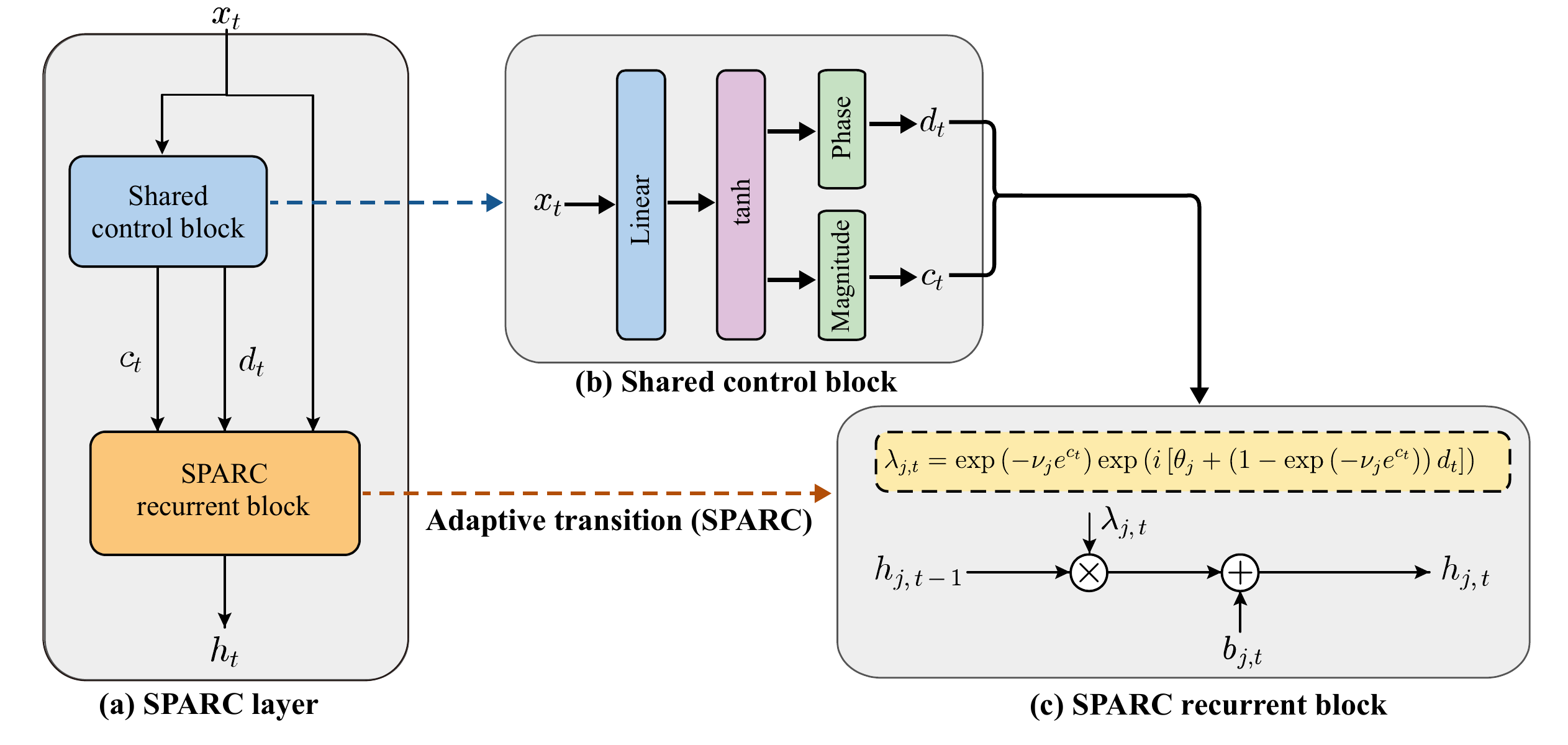}
    \caption{\textbf{Overview of SPARC.}
    (a) The layer input feeds both the shared control heads and the modal recurrence.
    (b) Two scalar heads produce the shared phase control $d_t$ and retention control $c_t$ from the current input.
    (c) Mode $j$ multiplies its previous state by an input-dependent complex transition and adds the current write contribution.}
    \label{fig:method_overview}
\end{figure}

SPARC combines mode-specific decay rates and frequencies with two shared retention and phase signals. Figure~\ref{fig:method_overview} shows the controls and mode-wise update, whose affine structure supports parallel computation and exact structured RTRL.

\subsection{Spectral Recurrence Preliminaries}
\label{sec:spectral_preliminaries}

Linear recurrent models compress history into a fixed-dimensional state. Consider
\begin{equation}
    \mathbf{h}_t=\mathbf{A}\mathbf{h}_{t-1}+\mathbf{B}\mathbf{x}_t,
    \label{eq:linear_recurrence}
\end{equation}
where $\mathbf{x}_t\in\mathbb{R}^{D}$, $\mathbf{h}_t\in\mathbb{C}^{H}$, $\mathbf{A}\in\mathbb{C}^{H\times H}$, and $\mathbf{B}\in\mathbb{C}^{H\times D}$. The transition propagates existing information; the input projection writes new content. Expressing a diagonalizable transition in its eigenbasis and absorbing the basis change into $\mathbf{h}_t$ and $\mathbf{B}$ gives
\begin{equation}
    \mathbf{h}_t=\boldsymbol{\lambda}\odot\mathbf{h}_{t-1}
                 +\mathbf{B}\mathbf{x}_t.
    \label{eq:spectral_recurrence}
\end{equation}
In this basis, each mode evolves independently through a complex transition. A stable transition can be written as
\begin{equation}
    \lambda_j=\exp(-\nu_j+i\theta_j),\qquad \nu_j>0,
    \label{eq:fixed_spectral_transition}
\end{equation}
where $e^{-\nu_j}$ controls retention and $\theta_j$ controls rotation. After $k$ steps, a stored contribution is multiplied by $e^{-k\nu_j}e^{ik\theta_j}$: decay determines its remaining magnitude, while phase determines its orientation and hence its projection onto a readout. Distinct modal decays and frequencies therefore provide heterogeneous memory responses, as in LRU and the linear RTU core \citep{orvieto2023lru,elelimy2024rtu}.

A fixed spectrum applies the same retention and rotation at every step. With an input-dependent transition and any additive input write, two historical states receiving the same current input satisfy
\begin{equation}
    h_{j,t}-h'_{j,t}
    =\lambda_{j,t}(\mathbf{x}_t)(h_{j,t-1}-h'_{j,t-1}).
    \label{eq:history_difference}
\end{equation}
Because the additive write cancels, gating the write alone cannot adapt how differences in stored history contract or rotate; selective evolution of stored memory requires an input-dependent transition. Appendix~\ref{app:write_transition_separation} relates this observation to accumulated phase.

\subsection{Shared Adaptive Spectral Dynamics}
\label{sec:shared_adaptive_dynamics}

Two bounded scalar heads with weights $\mathbf{w}_c,\mathbf{w}_d\in\mathbb{R}^{D}$ and biases $b_c,b_d\in\mathbb{R}$ generate the shared controls,
\begin{equation}
    c_t=\tanh(\mathbf{w}_c^\top\mathbf{x}_t+b_c),\qquad
    d_t=\tanh(\mathbf{w}_d^\top\mathbf{x}_t+b_d).
    \label{eq:shared_controllers}
\end{equation}
Each mode retains its learned decay $\nu_j>0$ and frequency $\theta_j$, with transition
\begin{equation}
    \rho_{j,t}=\exp(-\nu_j e^{c_t}),\qquad
    \lambda_{j,t}=\rho_{j,t}\exp\!\bigl(i[\theta_j+(1-\rho_{j,t})d_t]\bigr).
    \label{eq:sparc_transition}
\end{equation}
The retention control rescales modal decay rates together; the phase control adjusts their rotation. The factor $1-\rho_{j,t}$ reduces phase adjustments for persistent modes. Every transition remains strictly contractive because $|\lambda_{j,t}|=\rho_{j,t}<1$.

Sharing preserves relative decay rates: $(-\log\rho_{j,t})/(-\log\rho_{k,t})=\nu_j/\nu_k$. Thus $c_t$ changes the pace of forgetting while preserving spectral diversity. Through $1-\rho_{j,t}$, the shared phase signal also gives persistent modes smaller corrections than transient modes.

For the propagation of existing memory, fix the retention rates and baseline phases, and perturb each $d_t$ by at most $\epsilon$. Suppressing the mode index, let $P=\prod_{t=1}^{T}\lambda_t$ and $L=-\sum_{t=1}^{T}\log\rho_t$. The perturbed product $\widetilde P=\prod_{t=1}^{T}\widetilde\lambda_t$ obeys
\begin{equation}
    |\widetilde P-P|\leq\epsilon L e^{-L}\leq\epsilon/e.
    \label{eq:phase_clock_perturbation}
\end{equation}
Unlike unclocked phase errors that can accumulate with $T$, this bound is independent of $T$ and vanishes as $L\to0$, protecting persistent modes from phase-control perturbations. Appendix~\ref{app:phase_clock} gives the derivation.

The recurrence combines the transition with a normalized additive write:
\begin{equation}
    h_{j,t}=\lambda_{j,t}h_{j,t-1}
           +\sqrt{1-|\lambda_{j,t}|^2}\,b_{j,t}.
    \label{eq:sparc_recurrence}
\end{equation}
The input write $b_{j,t}$ applies real and imaginary content matrices with componentwise $\tanh$ activations, scaled by a learned gain and a modal write gate with mode-specific responses to $(c_t,d_t)$; Appendix~\ref{app:rtrl_derivation} gives the complete write map and its derivatives. Both transition and write depend only on $\mathbf{x}_t$, preserving affine dependence on $\mathbf{h}_{t-1}$, and the complex state $\mathbf{h}_t\in\mathbb{C}^{H}$ ($2H$ real coordinates) feeds a linear readout.

\paragraph{Connection to fixed spectral recurrences.}
\label{sec:fixed_spectral_limit}
Zero-initialized controllers give $c_t=d_t=0$, recovering the fixed spectrum in Eq.~\eqref{eq:fixed_spectral_transition}. With linear content and matched effective write parameters, this has the recurrent form of LRU; the corresponding decay-dependent write scale also recovers the linear RTU core \citep{orvieto2023lru,elelimy2024rtu}.

\subsection{Parallel BPTT and Structured RTRL}
\label{sec:parallel_sequence}
\label{sec:structured_online_sensitivities}
\label{sec:complexity}

Transition and write coefficients can be generated before recurrent propagation. Composing two steps of Eq.~\eqref{eq:sparc_recurrence} gives
\begin{equation}
    \mathbf{h}_2 = (\boldsymbol\lambda_2\odot\boldsymbol\lambda_1)\odot\mathbf{h}_0 +\boldsymbol\lambda_2\odot\sqrt{\mathbf{1}-|\boldsymbol\lambda_1|^2}\odot\mathbf{b}_1 +\sqrt{\mathbf{1}-|\boldsymbol\lambda_2|^2}\odot\mathbf{b}_2.
    \label{eq:affine_composition}
\end{equation}
Products, magnitudes, and square roots are elementwise. Affine composition is associative, so a work-efficient prefix scan evaluates the recurrent states with $O(TH)$ work and $O(\log T)$ temporal depth for a length-$T$ sequence \citep{blelloch1990prefix,martin2018parallel}.

Backpropagating through coefficient generation and the scan gives full-sequence BPTT. For online interaction, RTRL maintains state--parameter sensitivities forward through time and contracts them with the current loss gradient as observations arrive \citep{williams1989rtrl,elelimy2024rtu}.

For a recurrent parameter block $\boldsymbol{\psi}$, define the complex row Jacobian $Z^{\boldsymbol{\psi}}_{j,t}=\partial h_{j,t}/\partial\boldsymbol{\psi}$. Conditioned on the layer inputs $\mathbf{x}_{1:t}$, the chain rule gives
\begin{equation}
    Z^{\boldsymbol\psi}_{j,t} =\lambda_{j,t}Z^{\boldsymbol\psi}_{j,t-1} +h_{j,t-1}\partial_{\boldsymbol\psi}\lambda_{j,t} +\sqrt{1-|\lambda_{j,t}|^2}\,\partial_{\boldsymbol\psi}b_{j,t} -\frac{b_{j,t}}{\sqrt{1-|\lambda_{j,t}|^2}} \operatorname{Re}\!\left( \overline{\lambda}_{j,t}\partial_{\boldsymbol\psi}\lambda_{j,t}\right).
    \label{eq:rtrl_recurrence}
\end{equation}
Starting from $Z^{\boldsymbol{\psi}}_{j,0}=0$, these updates give exact sensitivities for recurrent-layer parameters. Modal parameters affect only their corresponding modes, while shared-controller derivatives reuse the scalar-head gradients $\partial_{\boldsymbol{\psi}}c_t$ and $\partial_{\boldsymbol{\psi}}d_t$ across modes. Following \citet{elelimy2024rtu}, upstream encoders use instantaneous gradients, and PPO updates reuse rollout-time traces. Appendix~\ref{app:rtrl_scope} describes this training procedure; Appendix~\ref{app:rtrl_derivation} derives the local updates and their contraction with readout gradients.

\begin{table}[t]
\centering
\caption{Partially observable continuous control.}
\label{tab:continuous_control}
\normalsize
\tabstyle\setlength{\tabcolsep}{3pt}
\begin{tabular}{@{}p{.17\linewidth}C{.15\linewidth}*{4}{C{\dimexpr(.68\linewidth-10\tabcolsep)/4\relax}}@{}}
\toprule
\textbf{Method} & \textbf{Training} & \textbf{Ant-P} & \textbf{Walker-P} & \textbf{Hopper-P} & \textbf{Cheetah-P}\\
\midrule
RTU & RTRL & {\boldmath\bfseries\tabmean{4757.32}{65.00}} & \tabmean{928.70}{151.41} & \tabmean{1369.88}{185.17} & \tabmean{2855.31}{24.50}\\
LRU & RTRL & \tabmean{4593.59}{759.21} & \tabmean{814.23}{98.81} & \tabmean{\underline{1376.24}}{84.58} & {\boldmath\bfseries\tabmean{2929.07}{6.58}}\\
RG-LRU & RTRL & \tabmean{2836.39}{502.32} & \tabmean{725.97}{174.55} & \tabmean{976.57}{161.77} & \tabmean{2806.78}{13.65}\\
GateLoop & RTRL & \tabmean{3807.76}{157.99} & \tabmean{\underline{1033.30}}{58.62} & \tabmean{1336.70}{62.33} & \tabmean{2638.09}{23.56}\\
Mamba-3 & TBPTT-16 & \tabmean{2143.40}{33.33} & \tabmean{784.51}{110.88} & \tabmean{1117.52}{470.56} & \tabmean{2867.58}{492.63}\\
Mamba-3 & RTRL & \tabmean{1803.42}{142.19} & \tabmean{564.54}{54.32} & \tabmean{846.86}{241.49} & \tabmean{2496.70}{252.47}\\
GRU & TBPTT-16 & \tabmean{4230.12}{432.69} & \tabmean{957.05}{194.44} & \tabmean{1128.68}{581.50} & \tabmean{2615.44}{80.67}\\
\tabgroup
\textbf{SPARC} & RTRL & \tabmean{\underline{4665.31}}{457.85} & {\boldmath\bfseries\tabmean{1153.70}{322.29}} & {\boldmath\bfseries\tabmean{1378.90}{139.94}} & \tabmean{\underline{2868.21}}{24.12}\\
\bottomrule
\end{tabular}
\tabnote{Mean return $\pm$ sample SD. Higher is better. Best means are bold; second-best means are underlined.}
\end{table}

The sparse dependency pattern makes these sensitivities compact. A row of either content matrix affects only its corresponding mode, as do its decay, frequency, gain, and write responses. Each shared controller affects all modes, but requires only one length-$(D+1)$ sensitivity vector per mode. A dense state-to-state Jacobian is therefore unnecessary.

Shared control uses $2(D+1)$ controller parameters and $O(D)$ projection work per step, versus $2H(D+1)$ and $O(HD)$ for independent dense controls per mode. For an $H$-dimensional state and $|\boldsymbol{\psi}|$ recurrent parameters, dense RTRL requires $O(H|\boldsymbol{\psi}|)$ trace storage and $O(H^2|\boldsymbol{\psi}|)$ work per step. SPARC's diagonal state Jacobian, mode-local parameters, and two shared controllers give $O(HD+H)$ running-trace storage and per-step work, including dense content projections. Appendix~\ref{app:complexity} details the control-path comparison, including RG-LRU's block-diagonal gates. Table~\ref{tab:rtrl_resources} reports RTRL resource measurements, and Section~\ref{sec:computational_efficiency} reports mixer latency and memory use.

\subsection{GPU-Efficient Implementation}
\label{sec:gpu_implementation}

Google DeepMind's released RecurrentGemma implementation provides a TPU-optimized Pallas scan and a reference PyTorch implementation \citep{botev2024recurrentgemma}. We implement GPU kernels for both RG-LRU and SPARC. SPARC divides time into chunks and modes into tiles. A prefix scan over local affine summaries supplies chunk entrance states, followed by parallel replay. The backward pass similarly uses a reverse scan for boundary cotangents, local adjoint replay, and gradient reduction.

The fused write kernel combines content activation, retention-dependent normalization, and modal gating. Rather than materializing $B\times T\times H$ transition tensors in global memory, the scan stores two $B\times T$ control arrays, where $B$ is batch size, and reconstructs modal transitions on-chip during propagation. Modal control gradients are reduced within each tile; recurrent accumulation and scan summaries use FP32. Appendix~\ref{app:gpu_implementation_details} provides the pseudocode (Algorithm~\ref{alg:chunked_scan}), memory layout, reset handling, and single-step kernel.

\section{Experiments}
\label{sec:experiments}

We evaluate SPARC under online reinforcement learning and full-sequence BPTT across control, memory, classification, and language modeling tasks.

\subsection{Experimental Setup}
\label{sec:experimental_setup}

\paragraph{Tasks and baselines.}
We evaluate four partially observable continuous-control tasks, six POPGym tasks \citep{morad2023popgym}, and five scalar-sequence classification tasks from UCR \citep{dau2019ucr}, MNIST, and CIFAR-10. Baselines are RTU, LRU, RG-LRU, GateLoop, Mamba-3, and GRU. RL state budgets are matched at $384$ real coordinates for continuous control and $128$ for POPGym.

\paragraph{Training and evaluation.}
RL uses the RTU actor-critic backbone and PPO \citep{elelimy2024rtu,schulman2017ppo}, with approximately 5M environment steps for continuous control and 4M for POPGym. SPARC, RTU, LRU, RG-LRU, and GateLoop use stored RTRL; GRU and Mamba-3 use TBPTT-16, alongside an additional Mamba-3 RTRL comparison. Classification uses a four-layer, width-$64$ residual backbone \citep{orvieto2023lru} and full-sequence BPTT for $20{,}000$ updates on the official training and test splits. For continuous control, POPGym, and the five classification tasks, we report the mean and sample SD over three seeded runs per method and task. Appendix~\ref{app:experimental_details} gives architectures, budgets, and evaluation procedures.

\subsection{Partially Observable Reinforcement Learning}
\label{sec:continuous_control}
\label{sec:popgym}

Inferring velocity from position-only observations requires smooth integration paired with rapid updates at contact changes. In Table~\ref{tab:continuous_control}, SPARC exceeds RG-LRU on all four continuous-control tasks and achieves the highest means on Walker-P and Hopper-P. Fixed spectral models remain competitive on regular gaits, with RTU leading Ant-P and LRU leading Cheetah-P, whereas unclocked real gates in RG-LRU and truncated windows in Mamba-3 degrade on Ant-P and Hopper-P. SPARC retains fixed-spectrum stability while adapting to contact dynamics on Walker-P.

\begin{figure*}[t]
 \centering
 \includegraphics[width=\textwidth]{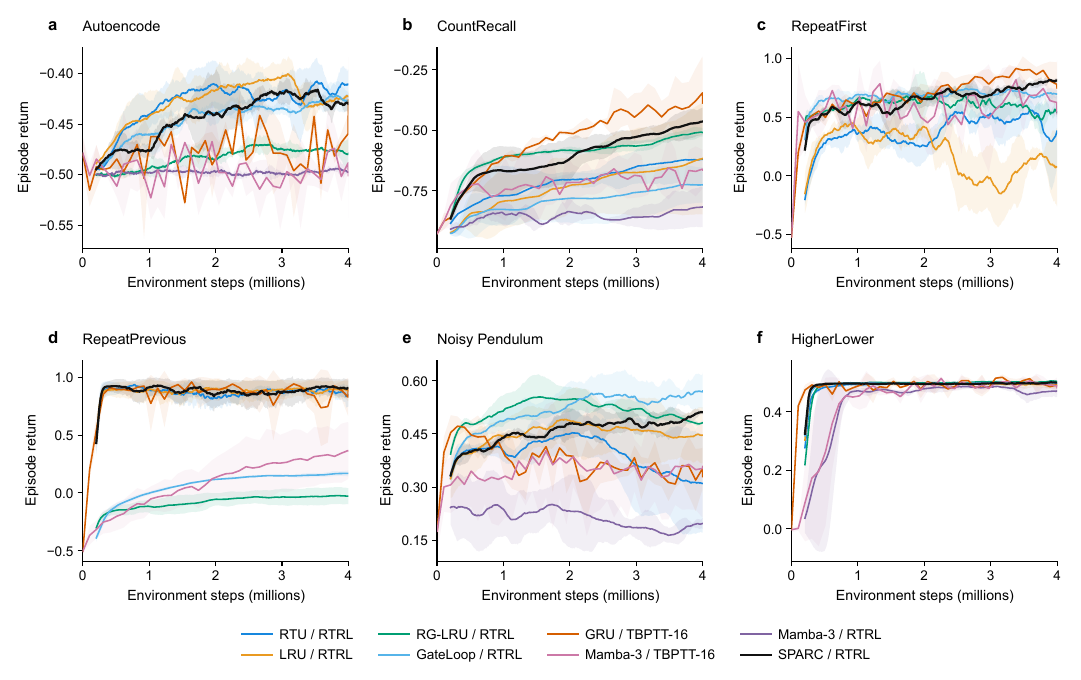}
 \caption{POPGym training curves across six partially observable tasks. Lines and shading show the mean return and one sample standard deviation across runs. RTRL returns are smoothed over $100$ rollouts; TBPTT returns are logged at coarser training intervals.}
 \label{fig:popgym_curves}
\end{figure*}

POPGym highlights a trade-off between selective gating and persistent propagation in Figure~\ref{fig:popgym_curves} and Table~\ref{tab:popgym}. Fixed spectra in RTU and LRU handle fixed-lag shifts well on RepeatPrevious but struggle on RepeatFirst, where later inputs overwrite the initial observation. Conversely, coordinate gates in RG-LRU and GateLoop improve RepeatFirst yet suffer cumulative drift on RepeatPrevious. By shielding slow modes with the phase clock while gating writes, SPARC converges reliably across both regimes, placing in the top two on all six tasks and leading Autoencode and RepeatPrevious.

\subsection{Sequence Modeling under Full-Sequence BPTT}
\label{sec:sequence_bptt}

Table~\ref{tab:sequence_classification} shows a similar pattern under full-sequence BPTT. On permuted MNIST and sequential CIFAR-10, where spatial structure spans $784$ to $1024$ scalar steps, RG-LRU, GateLoop, and Mamba-3 trail spectral models by over ten percentage points, reflecting the cost of unclocked coordinate gating on global phase coherence. SPARC preserves long-range spectral fidelity while adding clocked adaptation, leading permuted MNIST at $96.67\%$ and sequential CIFAR-10 at $63.11\%$ versus $62.28\%$ for GRU. Mamba-3 and RG-LRU instead lead UWave, where local gesture kinematics favor coordinate-level gating, while FordA and StarLightCurves show smaller gaps among top methods.

\begin{table}[t]
\centering
\caption{Test accuracy under full-sequence BPTT.}
\label{tab:sequence_classification}
\normalsize
\tabstyle
\begin{tabular}{@{}p{.16\linewidth}*{5}{C{\dimexpr(\linewidth-.16\linewidth-10\tabcolsep)/5\relax}}@{}}
\toprule
\textbf{Method} & \textbf{FordA} & \textbf{SLC} & \textbf{UWave} & \textbf{pMNIST} & \textbf{CIFAR-10}\\
\midrule
RTU & {\boldmath\bfseries\tabmean{94.70}{0.00}} & \tabmean{97.02}{0.15} & \tabmean{94.60}{1.04} & \tabmean{\underline{96.47}}{0.19} & \tabmean{61.36}{1.29}\\
LRU & \tabmean{\underline{94.66}}{0.80} & \tabmean{96.89}{0.26} & \tabmean{94.12}{1.15} & \tabmean{96.44}{0.20} & \tabmean{61.54}{1.38}\\
RG-LRU & \tabmean{93.03}{0.11} & \tabmean{96.92}{0.23} & \tabmean{\underline{96.35}}{0.06} & \tabmean{86.54}{0.23} & \tabmean{52.45}{1.08}\\
GateLoop & \tabmean{92.39}{0.05} & \tabmean{96.93}{0.17} & \tabmean{90.77}{0.58} & \tabmean{64.31}{2.40} & \tabmean{51.85}{0.34}\\
Mamba-3 & \tabmean{90.72}{1.66} & \tabmean{95.90}{0.26} & {\boldmath\bfseries\tabmean{96.57}{0.47}} & \tabmean{85.62}{4.93} & \tabmean{47.20}{2.36}\\
GRU & \tabmean{93.56}{0.21} & {\boldmath\bfseries\tabmean{97.20}{0.05}} & \tabmean{91.65}{0.35} & \tabmean{90.58}{0.56} & \tabmean{\underline{62.28}}{0.64}\\
\tabgroup
\textbf{SPARC} & \tabmean{\underline{94.66}}{0.16} & \tabmean{\underline{97.14}}{0.03} & \tabmean{94.29}{0.83} & {\boldmath\bfseries\tabmean{96.67}{0.29}} & {\boldmath\bfseries\tabmean{63.11}{1.26}}\\
\bottomrule
\end{tabular}
\tabnote{Mean test accuracy (\%) $\pm$ sample SD. SLC: StarLightCurves; pMNIST: permuted MNIST; CIFAR-10: sequential grayscale inputs. Best means are bold; second-best means are underlined.}
\end{table}

On Long Range Arena \citep{tay2020lra}, detailed in Appendix~\ref{app:lra_results}, SPARC exceeds all five spectral and selective baselines on Image, ListOps, Text, and Retrieval. GRU remains stronger on ListOps and Text, where sequential nonlinear state updates aid hierarchical and byte-level composition, whereas SPARC's clocked complex modes better preserve long-range spatial and matching signals on Image and Retrieval while retaining parallel scans.

\subsection{Language Modeling}
\label{sec:language_modeling}

We compare SPARC and RG-LRU in eight-layer language models of width $D=2048$, trained on approximately 5B FineWeb-Edu tokens \citep{lozhkov2024finewebedu} with the GPT-2 tokenizer, length-$2048$ sequences, and batch size $128$; Appendix~\ref{app:language_modeling} details the setup. On the 10M-token test set, SPARC lowers perplexity from $22.588$ to $22.136$, a $2.00\%$ reduction over RG-LRU.

\begin{figure*}[t]
\centering
\includegraphics[width=\textwidth]{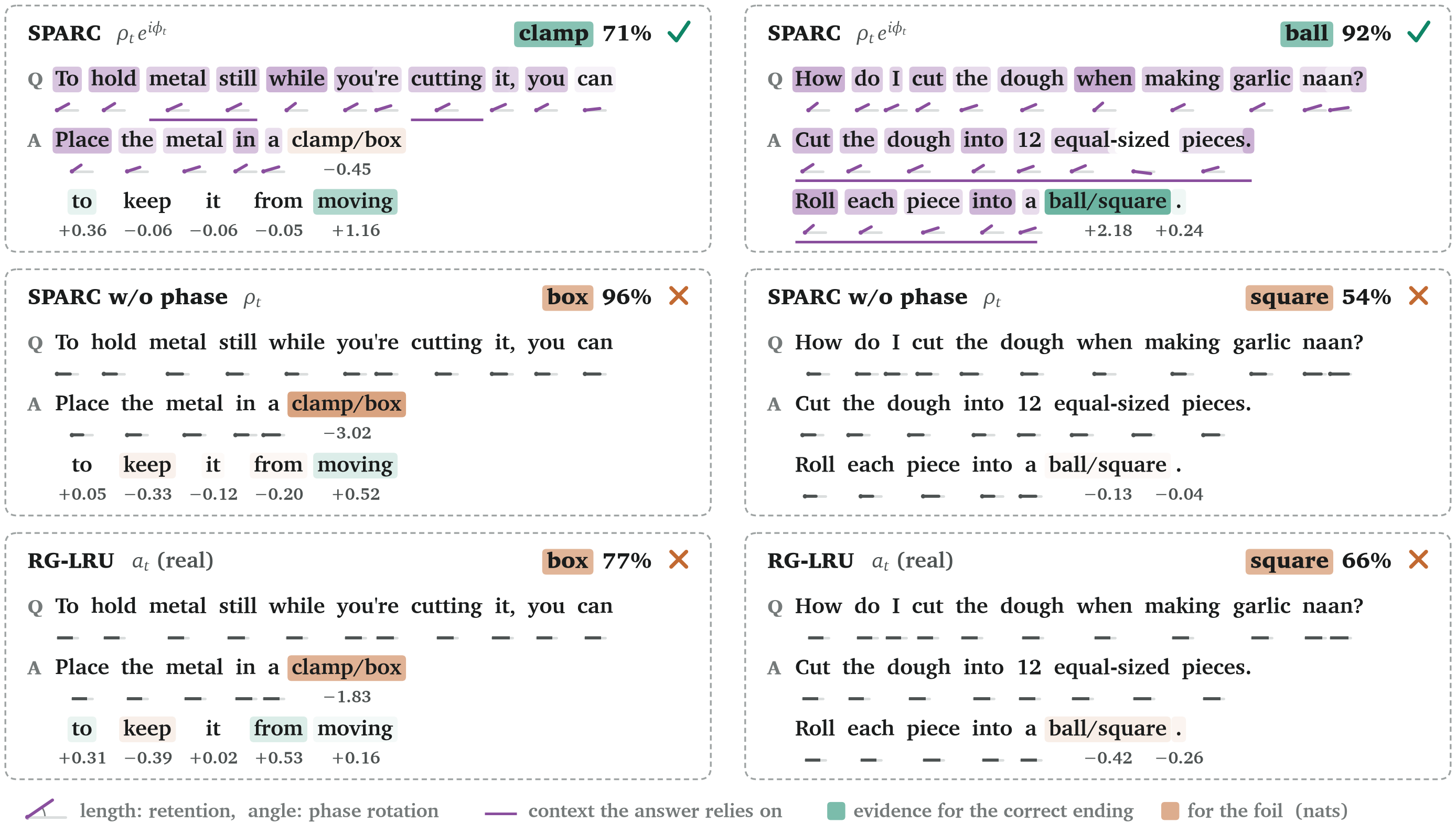}
\caption{\textbf{Adaptive phase shifts answer preferences on PIQA.} SPARC favors clamp at $71\%$ and ball at $92\%$. Ablating adaptive phase across all layers on the left, or layers 2 and 4 on the right, reverses both predictions. Percentages indicate normalized candidate likelihoods, and values show log-likelihood margins in nats. Underlines mark physical cues, and baseline rotations are retained throughout.}
\label{fig:language_model_examples}
\end{figure*}

Zero-shot results in Appendix~\ref{app:language_modeling} (Table~\ref{tab:zero_shot}) show higher accuracy on seven of eight tasks, including gains of $5.96$ percentage points on BoolQ, $2.53$ on RTE, and $2.07$ on PIQA, while WinoGrande favors RG-LRU. Figure~\ref{fig:language_model_examples} shows phase ablations reversing both PIQA preferences (clamp at $71\%$ and ball at $92\%$), with baseline rotations $\theta_j$ retained.

\subsection{Ablation Studies}
\label{sec:ablation}

\begin{table}[!htbp]
\centering
\caption{Core component ablations on classification and reinforcement learning tasks.}
\label{tab:phase_clock_main}
\normalsize
\tabstyle
\renewcommand{\arraystretch}{1.35}
\setlength{\tabcolsep}{3pt}
\begin{tabular}{@{}p{.22\linewidth}*{5}{C{\dimexpr(\linewidth-.22\linewidth-10\tabcolsep)/5\relax}}@{}}
\toprule
& \multicolumn{2}{c}{Classification (\%)} & \multicolumn{2}{c}{Continuous control} & POPGym\\
\cmidrule(lr){2-3}\cmidrule(lr){4-5}\cmidrule(l){6-6}
\textbf{Variant} & \textbf{SLC} & \textbf{sCIFAR-10} & \textbf{Walker-P} & \textbf{Cheetah-P} & \textbf{RepeatFirst}\\
\midrule
\textbf{SPARC} & {\boldmath\bfseries\tabmean{97.14}{0.03}} & {\boldmath\bfseries\tabmean{63.11}{1.26}} & {\boldmath\bfseries\tabmean{1153.70}{322.29}} & {\boldmath\bfseries\tabmean{2868.21}{24.12}} & \tabmean{\underline{0.817}}{0.045}\\
\tabgroup
Retention only & \tabmean{\underline{97.01}}{0.03} & \tabmean{61.54}{1.06} & \tabmean{894.66}{61.30} & \tabmean{2481.20}{19.50} & \tabmean{0.563}{0.282}\\
Phase only & \tabmean{96.65}{0.43} & \tabmean{61.42}{0.93} & \tabmean{715.22}{153.05} & \tabmean{2552.05}{0.63} & \tabmean{0.689}{0.093}\\
Static transition & \tabmean{96.75}{0.32} & \tabmean{61.04}{1.45} & \tabmean{893.84}{226.56} & \tabmean{2584.69}{101.78} & \tabmean{0.446}{0.353}\\
No phase clock & \tabmean{94.78}{1.08} & \tabmean{50.77}{3.38} & \tabmean{915.54}{260.21} & \tabmean{2541.74}{69.14} & \tabmean{0.350}{0.255}\\
Static phase clock & \tabmean{96.86}{0.37} & \tabmean{\underline{61.94}}{0.66} & \tabmean{\underline{923.81}}{196.03} & \tabmean{\underline{2618.26}}{8.33} & {\boldmath\bfseries\tabmean{0.866}{0.017}}\\
\bottomrule
\end{tabular}
\tabnote{Classification reports test accuracy (\%); RL reports returns. POPGym returns average the final 100 rollouts within each run.}
\end{table}

Table~\ref{tab:phase_clock_main} separates joint retention and phase adaptation from clock attenuation. Retaining only adaptive decay or only adaptive phase underperforms full SPARC across classification and control, indicating that magnitude and rotational updates are complementary. Removing the clock applies $d_t$ unattenuated to every mode, reducing sequential CIFAR-10 accuracy by $12.34$ percentage points and RepeatFirst return from $0.817$ to $0.350$ through rotational drift in persistent modes. A static clock $1-e^{-\nu_j}$ recovers most of the CIFAR-10 drop and leads RepeatFirst at $0.866$, where a single initial observation must remain protected. Coupling the clock to instantaneous retention $\rho_{j,t}$ further improves sequential CIFAR-10, Walker-P, and Cheetah-P, where new inputs repeatedly replace transient state components.

Appendix~\ref{app:measured_ablations} reports the full classification, continuous-control, and POPGym ablations, and Appendix~\ref{app:unshared_controls} compares control groups $G\in\{1,4,16,H\}$ at matched state capacity and training budget.

\paragraph{Control sharing.}
Increasing $G$ at matched state capacity gives no monotonic gain. Shared control at $G=1$ yields higher returns and lower cross-run variance than independent controls at $G=H$ on Ant-P and Hopper-P, consistent with low-dimensional control regularizing noisy policy updates. Independent controls still benefit UWave and Cheetah-P, yet mode-specific spectral and write parameters preserve heterogeneous responses under $G=1$ at much lower control cost.

\section{Computational Efficiency}
\label{sec:computational_efficiency}

\label{sec:efficiency_setup}
\begin{figure*}[t]
 \centering
 \includegraphics[width=\textwidth]{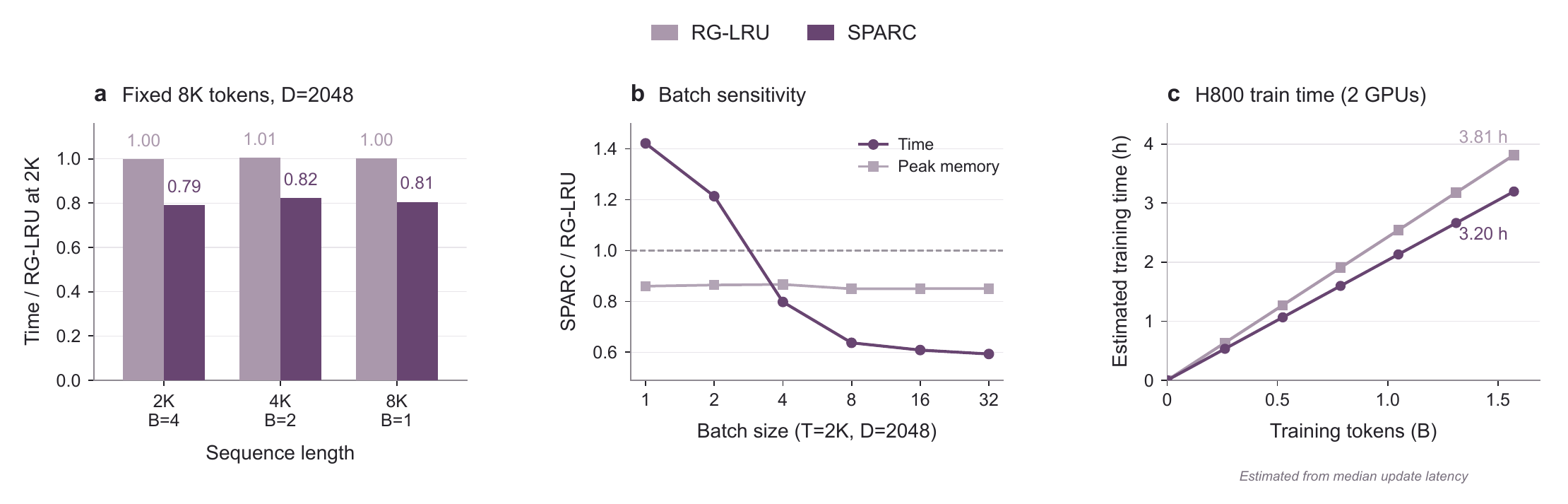}
 \caption{GPU efficiency of SPARC and RG-LRU. (a) Recurrent-mixer forward and backward timing at a fixed budget of 8K tokens and width $D=2048$. (b) Runtime and peak allocated-memory ratios, SPARC divided by RG-LRU, as batch size varies at $T=2048$, $D=2048$. (c) Estimated training time on two H800 GPUs from median update latency. Ratios below one favor SPARC.}
 \label{fig:gpu_efficiency}
\end{figure*}

We benchmark SPARC and our GPU-optimized RG-LRU kernels on an RTX PRO 6000 GPU in Figure~\ref{fig:gpu_efficiency}(a,b). Measurements cover coefficient generation and complete mixer forward and backward passes at matched workloads, with BF16 inputs and outputs and FP32 recurrent accumulation; Appendix~\ref{app:efficiency_protocol} gives the protocol.

\label{sec:efficiency_results}
\label{sec:scan_efficiency}
\label{sec:training_latency}
\label{sec:batch_memory}
At $BT=8192$, SPARC reduces latency across all three sequence lengths in Figure~\ref{fig:gpu_efficiency}(a) by reconstructing transitions from two $B\times T$ control streams on-chip. At $B=1,2$, multi-stage scan launches make SPARC $42.2\%$ and $21.4\%$ slower, respectively. It surpasses RG-LRU at $B=4$ and reaches a $1.69\times$ speedup at $B=32$ with $13\%$ to $15\%$ lower peak memory (Figure~\ref{fig:gpu_efficiency}(b)). Two-H800 language-model updates are $1.19\times$ faster (Figure~\ref{fig:gpu_efficiency}(c); Table~\ref{tab:language_model_timing}). Appendix~\ref{app:naive_performance} compares SPARC against sequential PyTorch, and Appendix~\ref{app:naive_correctness} verifies numerical agreement.

Comparing $G\in\{1,4,16,H\}$ within one kernel pipeline on an A800 GPU isolates control sharing in Appendix~\ref{app:control_cost_isolation}. Relative to independent controls at $G=H$, shared control at $G=1$ reduces latency by $8.79\%$ and peak memory by $28.64\%$ for $D=H=64$, expanding to a $5.41\times$ speedup and $37.03\%$ lower memory at $D=2048$, $H=1024$, where $O(HD)$ control projections and $2BTH$ intermediate streams become costly.

\section{Discussion}
\label{sec:discussion}

The grouping experiments distinguish memory width from control dimension: modal spectra and write responses preserve temporal diversity under two shared controls. Independent control improves UWave by $0.63$ percentage points and permuted MNIST by $0.08$, while sharing yields higher means and smaller cross-run variation on Ant-P and Hopper-P. The phase-clock study similarly separates attenuation from dynamic coupling. A static clock is competitive on permuted MNIST and protects RepeatFirst's initial observation. Dynamic coupling raises sequential CIFAR-10 accuracy from $61.94\%$ to $63.11\%$: the factor $1-\rho_{j,t}$ allows larger phase corrections when new inputs replace stored content while protecting persistent modes.

\section{Conclusion}
\label{sec:conclusion}

SPARC modulates heterogeneous complex memory through two shared retention and phase scalars. A retention-coupled phase clock protects persistent modes while allowing transient modes to rotate; modal gates regulate writes from the same signals. The diagonal affine recurrence supports parallel chunked scans with on-chip transition reconstruction and $O(HD+H)$ exact sensitivities for online RTRL. Across partially observable reinforcement learning, sequence classification, and language modeling, SPARC combines selective filtering with long-horizon recall and efficient training. Future work includes hybrid sequence architectures and structured online credit assignment across deeper networks.

\clearpage
\section*{AI Use Statement}
Generative AI tools were used solely to assist with code development, LaTeX typesetting, and language polishing. All mathematical derivations, implementations, experimental results, and text were verified by the authors, who take full responsibility for the content of this work.
\linespread{1}\selectfont
\bibliographystyle{iclr2027_conference}
\bibliography{references}

\clearpage
\appendix
\flushbottom
\setlength{\parskip}{3pt plus 1pt minus 1pt}
\setlength{\intextsep}{8pt plus 2pt minus 2pt}
\setlength{\textfloatsep}{10pt plus 2pt minus 2pt}
\setlength{\floatsep}{8pt plus 2pt minus 2pt}
\setlength{\abovecaptionskip}{4pt}
\setlength{\LTpre}{6pt}
\setlength{\LTpost}{6pt}
\setlength{\LTcapwidth}{\textwidth}
\makeatletter
\setlength{\@fptop}{0pt}
\setlength{\@fpsep}{18pt}
\makeatother

\section{Writing and Transition Control}
\label{app:write_transition_separation}

\subsection{Comparison with selective recurrences}
\label{app:selective_mechanisms}

Table~\ref{tab:selective_mechanisms} writes transition coordinates as $\lambda_{j,t}=\rho_{j,t}e^{i\phi_{j,t}}$. Indices $d$ and $h$ denote channels and heads. The transition controls and input writes are listed separately; $q_{j,t}$ is a write gate and $f_j(\mathbf{x}_t)$ denotes a learned content map.

\begin{table}[H]
\centering
\caption{Control, transition, and write forms of selective recurrences.}
\label{tab:selective_mechanisms}
\small
\setlength{\tabcolsep}{3pt}
\renewcommand{\arraystretch}{1.14}
\begin{tabular}{@{}L{.24\linewidth}L{.23\linewidth}L{.21\linewidth}L{\dimexpr.32\linewidth-6\tabcolsep\relax}@{}}
\multicolumn{4}{@{}l}{\textbf{(a) Transition controls}}\\[3pt]
\toprule
Method & Control signals & Retention $\rho_{j,t}$ & Phase $\phi_{j,t}$\\
\midrule
Mamba & Per channel: $\Delta_{d,t}$ & $e^{-\nu_j\Delta_{d,t}}$ & $0$ (real);\newline$\theta_j\Delta_{d,t}$ (complex)\\
\addlinespace[3pt]
Mamba-2 / mLSTM & Per head: $a_{h,t}$ & $a_{h,t}$ & $0$\\
\addlinespace[3pt]
RWKV-6 / GLA & Low rank to $O(D)$ vector & $e^{-\nu_j e^{u_{j,t}}}$ / $g_{j,t}$ & $0$\\
\addlinespace[3pt]
RG-LRU / CG-LRU & Per mode: $r_{j,t}$ & $e^{-\nu_jr_{j,t}}$ & $0$ (RG-LRU);\newline$\theta_jr_{j,t}$ (CG-LRU)\\
\addlinespace[3pt]
GateLoop (diagonal) & Per mode: $\alpha_{j,t},\varphi_{j,t}$ & $\sigma(\alpha_{j,t})$ & $\varphi_{j,t}$\\
\addlinespace[3pt]
Mamba-3 & Head/channel parameterized & Selective real decay & Data-dependent rotation\\
\addlinespace[3pt]
\textbf{SPARC} & Two layer-wide\newline scalars $(c_t,d_t)$ & $e^{-\nu_j e^{c_t}}$ & $\theta_j+(1-\rho_{j,t})d_t$\\
\bottomrule
\end{tabular}

\vspace{7pt}
\begin{tabular}{@{}L{.24\linewidth}L{.44\linewidth}L{\dimexpr.32\linewidth-4\tabcolsep\relax}@{}}
\multicolumn{3}{@{}l}{\textbf{(b) Input writes and recurrent state}}\\[3pt]
\toprule
Method & Input write & Recurrent state\\
\midrule
Mamba & Discretized write; selective $\mathbf{B}_t,\mathbf{C}_t$ & Expanded channel state\\
\addlinespace[2pt]
Mamba-2 / mLSTM & Key--value outer products & Matrix state per head\\
\addlinespace[2pt]
RWKV-6 / GLA & Key--value outer products & Matrix state\\
\addlinespace[2pt]
RG-LRU / CG-LRU & $\sqrt{1-\rho_{j,t}^{2}}\,q_{j,t}f_j(\mathbf{x}_t)$ & Diagonal state\\
\addlinespace[2pt]
GateLoop (diagonal) & Input-dependent complex write & Diagonal complex state\\
\addlinespace[2pt]
Mamba-3 & Two-endpoint write & Recurrent caches\\
\addlinespace[2pt]
\textbf{SPARC} & $\sqrt{1-\rho_{j,t}^{2}}\,\tilde\sigma_j(c_t,d_t)\,f_j(\mathbf{x}_t)$ & Diagonal complex state\\
\bottomrule
\end{tabular}
\tabnote{The modal indices of channel-specific spectra are suppressed. Fixed RG-LRU/CG-LRU scaling constants are absorbed into $\nu_j,\theta_j$. In SPARC, $\tilde\sigma_j$ includes the normalized modal write factor in Eq.~\eqref{eq:implementation_write}. $\mathbf{B}_t,\mathbf{C}_t$ denote Mamba's selective input and output maps.}
\end{table}

With $\Delta_{d,t}=e^{c_t}$ and matched modal rates, Mamba's retained magnitude equals SPARC's. In complex Mamba and CG-LRU, the step size or recurrence gate scales both attenuation and rotation \citep{gu2023mamba,de2024griffin}. SPARC supplies a distinct phase signal $d_t$ and scales its correction by $1-\rho_{j,t}$, preserving the baseline frequency $\theta_j$.

Mamba-2 and mLSTM apply the same scalar attenuation to all state coordinates within a head \citep{dao2024mamba2,beck2024xlstm}. RWKV-6 and GLA use low-rank generators with vector-valued decay outputs \citep{peng2024eagle,yang2023gla}. SPARC retains distinct modal $\nu_j,\theta_j$ while communicating two $B\times T$ control arrays and reconstructing modal transitions inside the scan kernel. Table~\ref{tab:control_cost_wide} measures the resulting control-sharing costs under a common implementation.

GateLoop predicts per-mode magnitudes and phases directly \citep{katsch2023gateloop}; SPARC adds a separate modal spectral baseline and the retention-coupled phase clock. Mamba-3 combines input-dependent rotations with a two-endpoint write \citep{lahoti2026mamba3}. Its RoPE-based implementation maintains cumulative angles and previous key/value components alongside the SSM state. RTRL propagates sensitivities through this augmented recurrent state. SPARC's diagonal complex state and shared controls give $O(HD+H)$ running-trace storage; Table~\ref{tab:rtrl_resources} reports the measured resource costs.

\subsection{Writing and stored-state evolution}

If an adaptive transition $\mathbf{A}(\mathbf{x})$ and a fixed transition $\mathbf{A}_0$ produce the same update for every reachable state after adjusting the additive input write, subtracting two such updates gives
\begin{equation}
 [\mathbf{A}(\mathbf{x})-\mathbf{A}_0](\mathbf{h}-\mathbf{h}')=0.
 \label{eq:app_difference_span}
\end{equation}
Thus $\mathbf{A}(\mathbf{x})=\mathbf{A}_0$ whenever reachable-state differences span the state space: an additive write cannot replace a different transformation of stored history in the same representation.

\paragraph{Accumulated phase.}
\label{app:phase_interpretation}

Expanding Eq.~\eqref{eq:sparc_recurrence} over a fixed input sequence yields
\begin{equation}
 h_{j,t} =\left(\prod_{\ell=1}^{t}\lambda_{j,\ell}\right)h_{j,0} +\sum_{k=1}^{t}\left(\prod_{\ell=k+1}^{t}\lambda_{j,\ell}\right) \sqrt{1-|\lambda_{j,k}|^2}\,b_{j,k}.
 \label{eq:app_phase_history_expansion}
\end{equation}
An empty product is one. A write from step $k$ is attenuated by the subsequent retention factors and rotated through the angle
\begin{equation}
 (t-k)\theta_j+\sum_{\ell=k+1}^{t}(1-\rho_{j,\ell})d_\ell.
 \label{eq:app_single_write}
\end{equation}
The baseline rotation depends on elapsed time; the shared phase signal adds an input-dependent correction. Accumulated phase changes a contribution's projection onto a readout at fixed retained magnitude. Relative phases also determine how past contributions reinforce or cancel.

\section{Structured RTRL Derivatives}
\label{app:rtrl_derivation}

\subsection{Input write}

The modal write combines real and imaginary content matrices $\mathbf{B}^{\mathrm{Re}},\mathbf{B}^{\mathrm{Im}}\in\mathbb{R}^{H\times D}$ with a normalized sigmoid gate:
\begin{equation}
 b_{j,t}=\frac{e^{\beta_j}\sigma(s_jd_t+v_jc_t)}
 {\sigma(0)\sqrt{1-|\lambda_j|^2}}
 \left[\tanh(\mathbf{B}^{\mathrm{Re}}_{j,:}\mathbf{x}_t)
       +i\tanh(\mathbf{B}^{\mathrm{Im}}_{j,:}\mathbf{x}_t)\right].
 \label{eq:implementation_write}
\end{equation}
Here $\sigma$ is the sigmoid, $\beta_j$ is a learned reference log gain, and $s_j,v_j$ give each mode its own response to the shared controls. The real and imaginary projections bound content with $\tanh$; the denominator uses the baseline $\lambda_j$ from Eq.~\eqref{eq:fixed_spectral_transition}.

At zero control, the sigmoid ratio equals one and the baseline denominator cancels the recurrence's write normalization. Thus $e^{\beta_j}$ sets the initial content amplitude independently of the modal decay. Away from this reference, retention changes the write scale and the modal gate selects how strongly each mode accepts the projected content. Shared control can consequently coordinate memory evolution and writing while their mode-specific responses remain distinct.

The recurrent parameters comprise the two content matrices, two shared controller weights and biases, and five modal vectors: decay rates $\nu_j$, frequencies $\theta_j$, reference log gains $\beta_j$, and response coefficients $s_j,v_j$.

\subsection{Local derivatives}

For the transition in Eq.~\eqref{eq:sparc_transition},
\begin{equation}
\begin{aligned}
 \frac{\partial\lambda_{j,t}}{\partial d_t}
 &=i(1-\rho_{j,t})\lambda_{j,t},\\
 \frac{\partial\lambda_{j,t}}{\partial c_t}
 &=\nu_j e^{c_t}(-1+i\rho_{j,t}d_t)\lambda_{j,t},\\
 \frac{\partial\lambda_{j,t}}{\partial\nu_j}
 &=e^{c_t}(-1+i\rho_{j,t}d_t)\lambda_{j,t},\\
 \frac{\partial\lambda_{j,t}}{\partial\theta_j}
 &=i\lambda_{j,t}.
\end{aligned}
\label{eq:app_transition_derivatives}
\end{equation}
The remaining recurrent parameters have zero direct transition derivative. Differentiating the input write directly gives
\begin{equation}
\begin{aligned}
 \frac{\partial b_{j,t}}{\partial d_t}
 &=b_{j,t}s_j\sigma(-s_jd_t-v_jc_t),\\
 \frac{\partial b_{j,t}}{\partial c_t}
 &=b_{j,t}v_j\sigma(-s_jd_t-v_jc_t),\\
 \frac{\partial b_{j,t}}{\partial\nu_j}
 &=-b_{j,t}\frac{|\lambda_j|^2}{1-|\lambda_j|^2},\\
 \frac{\partial b_{j,t}}{\partial\beta_j}
 &=b_{j,t},\qquad
 \frac{\partial b_{j,t}}{\partial\theta_j}=0,\\
 \frac{\partial b_{j,t}}{\partial s_j}
 &=b_{j,t}d_t\sigma(-s_jd_t-v_jc_t),\\
 \frac{\partial b_{j,t}}{\partial v_j}
 &=b_{j,t}c_t\sigma(-s_jd_t-v_jc_t).
\end{aligned}
\label{eq:app_write_derivatives}
\end{equation}
For log-coordinate optimization, multiply the $\nu_j$ and $\theta_j$ derivatives by $\nu_j$ and $\theta_j$, respectively. Content-matrix derivatives follow directly from Eq.~\eqref{eq:implementation_write}: differentiate the corresponding projected component using the activation derivative $1-\tanh^2(z)$, or one for linear content. Row $j$ directly affects mode $j$. Controller-weight derivatives follow the chain rule through the two shared scalar heads, including the fixed output-range scales in Appendix~\ref{app:control_parameterization}.

\subsection{Sensitivity propagation and episode resets}

Let $m_t$ be one within an episode and zero at its first step. The reset-aware recurrence is
\begin{equation}
 h_{j,t}=m_t\lambda_{j,t}h_{j,t-1}
         +\sqrt{1-|\lambda_{j,t}|^2}\,b_{j,t}.
 \label{eq:app_masked_scan}
\end{equation}
Differentiating this expression for a real parameter block $\boldsymbol\psi$ gives
\begin{equation}
\begin{aligned}
 Z^{\boldsymbol\psi}_{j,t}
 &=m_t\lambda_{j,t}Z^{\boldsymbol\psi}_{j,t-1}
  +m_t h_{j,t-1}\partial_{\boldsymbol\psi}\lambda_{j,t}\\
 &\quad+\sqrt{1-|\lambda_{j,t}|^2}\,
               \partial_{\boldsymbol\psi}b_{j,t}
 -\frac{b_{j,t}}{\sqrt{1-|\lambda_{j,t}|^2}}
       \operatorname{Re}\!\left(
        \overline{\lambda}_{j,t}\partial_{\boldsymbol\psi}\lambda_{j,t}
       \right).
\end{aligned}
\label{eq:app_reset_trace}
\end{equation}
The last term differentiates the retention-dependent normalization. The reset mask clears the historical state and sensitivity, and the current normalized write starts the new episode. With a parameter-independent initial state, the initial sensitivities are zero.

Zero-initialized controllers make the write gate initially neutral. Gradients reach the controllers through the transition and write pathways.

For a real loss $\mathcal{L}_t$, the readout supplies state gradients in real coordinates. Contracting them with the sensitivities gives
\begin{equation}
\begin{aligned}
 \frac{\partial\mathcal{L}_t}{\partial\boldsymbol\psi}
 =\sum_{j=1}^{H}\Bigl(&
 \frac{\partial\mathcal{L}_t}{\partial\operatorname{Re}h_{j,t}}
       \operatorname{Re}Z^{\boldsymbol\psi}_{j,t}
 +\frac{\partial\mathcal{L}_t}{\partial\operatorname{Im}h_{j,t}}
       \operatorname{Im}Z^{\boldsymbol\psi}_{j,t}\Bigr)
 +\left.\frac{\partial\mathcal{L}_t}{\partial\boldsymbol\psi}\right|_{\mathrm{direct}}.
\end{aligned}
\label{eq:online_loss_gradient}
\end{equation}
The direct term differentiates the loss at fixed recurrent state. The two content matrices, five modal vectors, and two shared controllers require $2HD+5H+2H(D+1)=4HD+7H$ complex trace entries. Trace storage and per-step updates are both $O(HD+H)$.

\subsection{RTRL--BPTT duality}
\label{app:rtrl_bptt_duality}

Let $F_j$ denote the reset-aware state update in Eq.~\eqref{eq:app_masked_scan}. Its local parameter Jacobian and the complex encoding of the readout gradient are
\begin{equation}
 \mathbf{g}^{\boldsymbol\psi}_{j,t} :=\left.\frac{\partial F_j(h_{j,t-1},\mathbf{x}_t;\boldsymbol\psi)} {\partial\boldsymbol\psi}\right|_{h_{j,t-1},\mathbf{x}_t}, \qquad u_{j,t} :=\frac{\partial\mathcal{L}_t}{\partial\operatorname{Re}h_{j,t}} +i\frac{\partial\mathcal{L}_t}{\partial\operatorname{Im}h_{j,t}}.
 \label{eq:app_local_jacobian_readout}
\end{equation}
Writing $a_{j,t}=m_t\lambda_{j,t}$, the forward sensitivity and backward adjoint recurrences are
\begin{equation}
 Z^{\boldsymbol\psi}_{j,t} =a_{j,t}Z^{\boldsymbol\psi}_{j,t-1} +\mathbf{g}^{\boldsymbol\psi}_{j,t}, \qquad \bar h_{j,k} =\overline{a}_{j,k+1}\bar h_{j,k+1}+u_{j,k}.
 \label{eq:app_rtrl_bptt_recursions}
\end{equation}
The forward recursion starts at $Z^{\boldsymbol\psi}_{j,0}=0$. The backward recursion applies for $1\leq k<T$, starting with $\bar h_{j,T}=u_{j,T}$ and zero terminal adjoint $\bar h_{j,T+1}=0$.

Expanding the two recurrences gives
\begin{equation}
 Z^{\boldsymbol\psi}_{j,t} =\sum_{k=1}^{t} \left(\prod_{\ell=k+1}^{t}a_{j,\ell}\right) \mathbf{g}^{\boldsymbol\psi}_{j,k}, \qquad \bar h_{j,k} =\sum_{t=k}^{T} \overline{\left(\prod_{\ell=k+1}^{t}a_{j,\ell}\right)}u_{j,t},
 \label{eq:app_rtrl_bptt_expansions}
\end{equation}
where an empty product is one. Interchanging the finite sums yields
\begin{equation}
 \sum_{t=1}^{T}\sum_{j=1}^{H}
 \operatorname{Re}\!\left(\overline{u_{j,t}}Z^{\boldsymbol\psi}_{j,t}\right)
 =\sum_{k=1}^{T}\sum_{j=1}^{H}
 \operatorname{Re}\!\left(\overline{\bar h_{j,k}}
                          \mathbf{g}^{\boldsymbol\psi}_{j,k}\right).
\label{eq:app_rtrl_bptt_duality}
\end{equation}
Adding $\sum_{t=1}^{T}\left.\partial\mathcal{L}_t/\partial\boldsymbol\psi\right|_{\mathrm{direct}}$ gives the same total recurrent-parameter gradient for $\mathcal{L}=\sum_{t=1}^{T}\mathcal{L}_t$. A reset makes every product crossing that boundary zero. BPTT traverses the length-$T$ state sequence backward, storing or recomputing it; RTRL updates the sensitivities forward and contracts them with the current readout gradient, using $O(HD+H)$ running-trace storage for SPARC.

\subsection{Stored sensitivities in PPO}
\label{app:rtrl_scope}

At fixed recurrent parameters and supplied layer inputs, Eq.~\eqref{eq:app_reset_trace} propagates the full recurrent-layer derivative through time. Temporal differentiation through an upstream encoder additionally requires propagating the input's parameter derivatives through the same rule. The RL implementation uses local encoder gradients.

Following RTU \citep{elelimy2024rtu}, recurrent sensitivities are evaluated during rollout collection and stored for PPO minibatches. They remain at their rollout-time values during subsequent parameter updates. Rollout storage uses $O(THD)$ memory, while the running trace state occupies $O(HD+H)$.

\subsection{Control-path complexity}
\label{app:complexity}

For input width $D$ and $H$ complex modes, independent retention and phase controls require $2H(D+1)$ parameters and $O(HD)$ projection work per step. Shared control uses $2(D+1)$ parameters and $O(D)$ work. Over a length-$T$ sequence, shared and independent outputs occupy $2T$ and $2HT$ values, respectively. Sharing also reduces backward projection work.

Modal coefficient generation costs $O(H)$ and dense content projection costs $O(HD)$, giving $O(HD+H)$ work per step. Structured RTRL has the same work and storage order as other diagonal cores with local parameter dependencies \citep{elelimy2024rtu}. RG-LRU reduces its projection cost with block-diagonal gates \citep{de2024griffin}. Section~\ref{sec:computational_efficiency} compares measured latency and memory use, including coefficient generation.

\section{Component Analysis and Controlled Diagnostics}
\label{app:ablations}

This section explains the components and defines the corresponding interventions. Full component ablation results across classification, continuous control, and POPGym are reported in Appendix~\ref{app:measured_ablations} (Tables~\ref{tab:component_main}--\ref{tab:core_popgym_clock}).

\subsection{Shared controls and retention-coupled phase}
\label{app:control_parameterization}

Concretely, our implementation scales the bounded $\tanh$ heads in Eq.~\eqref{eq:shared_controllers} with fixed output ranges, setting $c_t=\log(16)\tanh(\mathbf{w}_c^\top\mathbf{x}_t+b_c)$ and $d_t=\frac{\pi}{2}\tanh(\mathbf{w}_d^\top\mathbf{x}_t+b_d)$. The two heads have independent weights and biases; the modal write responses absorb these fixed scale factors. Decays $\nu_j$ and frequencies $\theta_j$ are optimized in log coordinates. The shared multiplier $e^{c_t}$ scales modal decay rates between $1/16$ and $16$ times their baselines while preserving their relative values.

\paragraph{Role of the phase clock.}
\label{app:phase_clock}
The factor $1-\rho_{j,t}$ attenuates phase corrections in persistent modes. For slowly decaying modes, $1-\rho_{j,t}$ is approximately $\nu_j e^{c_t}$, so the correction decreases with the decay. For rapidly forgotten modes, larger phase changes act on a previous-state contribution that is already attenuated. Removing this factor tests the retention dependence of phase adaptation.

Let $\widetilde d_t=d_t+\delta_t$, with $|\delta_t|\leq\epsilon$, while $\rho_t\in(0,1]$ and the baseline phase are fixed. For $P=\prod_{t=1}^{T}\lambda_t$, the accumulated phase change is $\Delta\Phi=\sum_{t=1}^{T}(1-\rho_t)\delta_t$, so $\widetilde P=P e^{i\Delta\Phi}$ and $|P|=e^{-L}$, where $L=-\sum_{t=1}^{T}\log\rho_t$. Hence
\begin{equation}
 |\widetilde P-P| =e^{-L}|e^{i\Delta\Phi}-1| \leq\epsilon e^{-L}\sum_{t=1}^{T}(1-\rho_t) \leq\epsilon L e^{-L}\leq\epsilon/e.
\end{equation}
The last two steps use $1-\rho_t\leq-\log\rho_t$ and $\sup_{L\geq0}Le^{-L}=1/e$. This bounds the transfer of an existing memory contribution uniformly in sequence length; for increasingly persistent modes, $L\to0$ and the bound vanishes.

\subsection{Content, write gate, normalization, and gain}
\label{app:write_design}

\paragraph{Input activation and saturation.}
Componentwise $\tanh$ bounds the real and imaginary projected inputs to $(-1,1)$, controlling the amplitude of newly written content. Its derivative $1-\tanh^2(z)$ decreases as the projected magnitude grows.

\paragraph{Modal write response.}
The normalized sigmoid factor in Eq.~\eqref{eq:implementation_write} gives each mode its own write response to the shared controls. Its value is one at zero control, preserving the reference write amplitude at initialization.

\paragraph{Retention-dependent normalization.}
The prefactor $\sqrt{1-|\lambda_{j,t}|^2}$ reduces the instantaneous write in persistent modes and increases it under stronger forgetting, following related recurrent normalizations \citep{orvieto2023lru,de2024griffin}. The baseline denominator separates this adaptive factor from the reference amplitude. Transition interventions preserve the adaptive write, so their comparison concerns how existing memory propagates.

\paragraph{Independent reference gain.}
The parameter $\beta_j$ sets a learned reference write scale independently of the decay rate $\nu_j$. With nonlinear content, it controls the amplitude after activation rather than changing the saturation threshold of the projected input.

\paragraph{Control sharing.}
Shared controllers use $2(D+1)$ parameters. Independent controllers per complex mode use $2H(D+1)$ parameters, increasing control-projection work from $O(D)$ to $O(HD)$ at the same $H$-mode state size. Running-trace storage is $O(HD+H)$ in both cases. Appendix~\ref{app:unshared_controls} compares $G=1,4,16,H$ at matched state capacity.

\subsection{Programmatically generated diagnostic tasks}
\label{app:controlled_diagnostics}

Figure~\ref{fig:motivation} and Table~\ref{tab:motivation_results} evaluate two memory requirements with generated sequences: preserving relevant events among distractors and tracking a state with changing dynamics.

\begin{table}[H]
\centering
\caption{Event-order recall and dynamical tracking.}
\label{tab:motivation_results}
\small
\tabstyle
\begin{tabular}{@{}p{.24\linewidth}*{2}{C{\dimexpr(\linewidth-.24\linewidth-4\tabcolsep)/2\relax}}@{}}
\toprule
\textbf{Method} & \shortstack{\textbf{Event-order}\\\textbf{error (\%)}} & \shortstack{\textbf{Tracking}\\\textbf{NMSE}}\\
\midrule
RTU & \tabmean{16.99}{0.91} & \tabmean{0.359}{0.004}\\
RG-LRU & {\boldmath\bfseries\tabmean{0.00}{0.00}} & \tabmean{\underline{0.345}}{0.007}\\
SPARC & {\boldmath\bfseries\tabmean{0.00}{0.00}} & {\boldmath\bfseries\tabmean{0.085}{0.002}}\\
\bottomrule
\end{tabular}
\tabnote{Mean $\pm$ sample SD over five seeds. Lower is better.}
\end{table}

\paragraph{Selective event-order retention.}
Each sequence contains three relevant events, labeled A, B, and C, among distractor inputs. The classification target is their order of occurrence: the recall schematic in Figure~\ref{fig:motivation} has target B--A--C. Recovering this order requires remembering the earlier event identities through the distractors and incorporating each later event in sequence. Classification error is the fraction of evaluation sequences with an incorrect order prediction. SPARC and RG-LRU both achieve zero error, compared with $16.99\%$ for RTU.

\paragraph{Event-driven dynamical tracking.}
The prediction target is a latent-state trajectory generated under event-dependent contraction and rotation. Event times mark changes in how the state evolves: contraction changes its retained magnitude, while rotation changes its direction and thus its coordinates. The model receives the generated input sequence and predicts the evolving state throughout the sequence. The right panel of Figure~\ref{fig:motivation} illustrates rotation and scaling as events change the dynamics. SPARC obtains NMSE $0.085$, compared with $0.359$ for RTU and $0.345$ for RG-LRU. NMSE is the summed squared prediction error divided by the summed squared target magnitude over the evaluation set.

\paragraph{Training and evaluation.}
Both diagnostics use length-$48$ sequences with four input channels, $384$ training sequences, and $512$ evaluation sequences. Models have $64$ complex modes or $128$ real state coordinates and a linear readout. Training uses AdamW with learning rate $0.01$, batch size $64$, no weight decay, and gradient-norm clipping at $1$. Event-order recall uses $1200$ updates and dynamical tracking $450$, with five training seeds per task.

\subsection{Matched component-removal definitions}

Table~\ref{tab:component_ablation_definitions} defines the reported transition interventions. Shared signals continue to drive the write pathway when their transition contribution is removed. Component results appear in Appendix~\ref{app:measured_ablations}; control grouping appears in Appendix~\ref{app:unshared_controls}.

\begin{table}[H]
\centering
\caption{Definitions of the reported component and control-sharing variants.}
\label{tab:component_ablation_definitions}
\small
\tabstyle
\begin{tabular}{@{}p{.25\linewidth}L{\dimexpr.75\linewidth-2\tabcolsep\relax}@{}}
\toprule
\textbf{Variant} & \textbf{Intervention}\\
\midrule
Retention only & Set $\lambda_{j,t}=\rho_{j,t}e^{i\theta_j}$; retain $d_t$ in the write gate.\\
Phase only & Set $\rho_{j,t}=|\lambda_j|$ only in the transition, including its phase clock; retain the original adaptive write normalization and gate.\\
No phase clock & Replace $(1-\rho_{j,t})d_t$ by $d_t$; keep retention and all write terms.\\
Static phase clock & Use $(1-e^{-\nu_j})d_t$; retain adaptive retention and writing.\\
Static transition & Set $\lambda_{j,t}=\lambda_j$ only in the transition; retain the original adaptive write normalization and gate.\\
Unshared control & Use independent retention and phase projections producing $c_{j,t},d_{j,t}$ for each mode.\\
\bottomrule
\end{tabular}
\end{table}

\section{Experimental Details and Hyperparameters}
\label{app:experimental_details}

Continuous-control, POPGym, and the five sequence-classification results each aggregate three independently seeded training runs per method and task. The reported mean and sample SD summarize these three runs.

\subsection{Reinforcement-learning architecture and optimization}
\label{app:rl_details}

The observation encoder, policy head, and value head each have width $64$. The recurrent input is the encoded current observation,
\begin{equation}
 \mathbf{x}_t=\tanh(\mathbf{W}_{\mathrm{enc}}\mathbf{o}_t+\mathbf{b}_{\mathrm{enc}}).
 \label{eq:actor_critic_backbone}
\end{equation}
SPARC's real and imaginary states are concatenated and passed through ReLU before the two heads. Continuous policies use a diagonal Gaussian with learned state-independent log standard deviation and action clipping; discrete policies use categorical logits.

Continuous control applies the partial-observation mask and enables observation and reward normalization. POPGym uses native Easy environments with their task noise and unnormalized observations and rewards. Noisy Pendulum uses the continuous-action policy, while the other five tasks use categorical policies. Table~\ref{tab:rl_hyperparameters} summarizes the shared reinforcement-learning hyperparameters.

\begin{table}[H]
\centering
\caption{Shared reinforcement-learning hyperparameters.}
\label{tab:rl_hyperparameters}
\small
\tabstyle
\begin{tabular}{@{}p{.50\linewidth}*{2}{C{\dimexpr(\linewidth-.50\linewidth-4\tabcolsep)/2\relax}}@{}}
\toprule
\textbf{Setting} & \textbf{Continuous control} & \textbf{POPGym}\\
\midrule
Environment interaction steps & $4{,}999{,}168$ & $3{,}999{,}744$\\
Physical rollout length & $2048$ & $2048$\\
PPO epochs per rollout & $4$ & $10$\\
Minibatches per epoch & $32$ & $32$\\
\tabgroup
Real-valued recurrent state coordinates & $384$ & $128$\\
SPARC complex modes & $192$ & $64$\\
Encoder width & $64$ & $64$\\
Policy-head / value-head width & $64/64$ & $64/64$\\
\tabgroup
Discount factor & $0.99$ & $0.99$\\
GAE parameter & $0.95$ & $0.95$\\
PPO clipping parameter & $0.2$ & $0.2$\\
Value-loss coefficient & $0.5$ & $0.5$\\
Entropy coefficient & $0$ & $0$\\
Global gradient-norm clipping & $0.5$ & $0.5$\\
Optimizer & Adam & Adam\\
Optimizer $\epsilon$ & $10^{-5}$ & $10^{-5}$\\
Observation normalization & Enabled & Disabled\\
Reward normalization & Enabled & Disabled\\
Previous action / reward inputs & Absent / absent & Absent / absent\\
RTRL minibatch sequence length & $1$ & $1$\\
GRU / Mamba-3 TBPTT length & $16$ & $16$\\
\bottomrule
\end{tabular}
\tabnote{Learning rates are fixed across methods within each task. The interaction budgets comprise $2441$ and $1953$ rollouts of $2048$ steps, respectively.}
\end{table}

The learning rate is $10^{-4}$ for all listed RL tasks except Walker-P ($3\times10^{-5}$), RepeatFirst ($10^{-3}$), and HigherLower ($10^{-5}$).

SPARC, RTU, LRU, RG-LRU, and GateLoop use stored recurrent sensitivities and minibatch sequence length $1$. GRU and Mamba-3 use TBPTT with length $16$. The additional Mamba-3 RTRL variant retains its core recurrence and recurrent caches, with the same state budget and common readout; its stored sensitivities include the additional recurrent state.

Both SPARC controllers use the full task learning rate. PPO updates reuse rollout-time value targets and cached sensitivity traces. Recurrent states and traces reset at episode boundaries. RL states use complex64, with single-precision real parameters and projections.

\subsection{Spectral and controller initialization}
\label{app:initialization}

\paragraph{Shared initialization.}
Controller vectors and biases are initialized to zero, giving $c_t=d_t=0$ and a neutral write gate. In classification and RL, the modal responses acting on the unscaled $\tanh$ outputs start at one, equivalent to $s_j=2/\pi$ and $v_j=1/\log16$ in Eq.~\eqref{eq:implementation_write}; language-model responses start at zero. The initial write is projected content scaled by $e^{\beta_j}$. Content projections have no bias, and recurrent states start at zero.

\paragraph{Reinforcement learning.}
SPARC and RTU share seed-paired initializations for the encoder, readout, modal spectrum, content projections, and reference gain. Modal radii follow the same area-uniform distribution, and frequencies follow the same baseline distribution. Log decay, log frequency, and reference log gain are optimized independently.

\paragraph{Classification.}
SPARC and LRU share seed-paired initializations for the spectrum, content projections, readout, and reference gain. The area-uniform annulus near the unit circle supplies slowly decaying modes for long-range propagation \citep{orvieto2023lru}.

\subsection{Full-sequence architecture and optimization}
\label{app:bptt_details}

Classification uses four unidirectional residual sequence layers of model width $64$, pre-layer normalization, the full-GLU activation block, dropout $0.1$, and sequence-mean pooling. SPARC has $64$ complex modes per layer. The learned complex readout matrix $\mathbf{C}$ and real skip coefficients $\mathbf{D}_{\mathrm{skip}}$ give
\begin{equation}
 \mathbf{y}_t=\operatorname{Re}(\mathbf{C}\mathbf{h}_t)
                 +\mathbf{D}_{\mathrm{skip}}\odot\mathbf{x}_t.
 \label{eq:app_bptt_readout}
\end{equation}
A scalar-input embedding feeds the stack; each residual layer contains the recurrent mixer, readout, and activation block. The readout and skip coefficients use the non-recurrent optimizer described below.

Training uses $20{,}000$ full-sequence updates at batch size $32$. The non-recurrent learning rate warms up from $10^{-6}$ to $1.95\times10^{-3}$ over $2000$ updates, then follows cosine decay to $10^{-6}$. The recurrent learning rate is one quarter of this rate throughout. Recurrent parameters use Adam \citep{kingma2015adam} with zero weight decay; non-recurrent parameters use AdamW \citep{loshchilov2019adamw} with weight decay $0.05$.

\subsection{Classification data and preprocessing}
\label{app:classification_data}

All five tasks use the official training and test splits. Inputs have one scalar channel; Table~\ref{tab:classification_data} gives the sequence lengths and class counts. Reported test accuracies use the checkpoint after $20{,}000$ training updates.

\begin{table}[H]
\centering
\caption{Classification data configurations.}
\label{tab:classification_data}
\small
\tabstyle
\begin{tabular}{@{}p{.43\linewidth}*{2}{C{\dimexpr(\linewidth-.43\linewidth-4\tabcolsep)/2\relax}}@{}}
\toprule
\textbf{Dataset} & \textbf{Length} & \textbf{Classes}\\
\midrule
FordA & $500$ & $2$\\
StarLightCurves & $1024$ & $3$\\
UWaveGestureLibraryAll & $945$ & $8$\\
Permuted MNIST & $784$ & $10$\\
Seq. CIFAR-10 Gray & $1024$ & $10$\\
\bottomrule
\end{tabular}
\tabnote{Length denotes the number of scalar sequence positions.}
\end{table}

UCR preprocessing uses a global z-score transformation fitted on the training data. Pixel normalization also uses training data, without augmentation. Permuted MNIST uses one fixed pixel permutation, and grayscale CIFAR-10 uses row-major ordering. Data splits and the permutation are fixed across models and training runs. Table~\ref{tab:sequence_classification} aggregates test accuracy across runs.

\subsection{Learning curves}
\label{app:metrics_figures}

RTRL returns are logged at every rollout and smoothed with a $100$-rollout moving average within each run. TBPTT training returns are logged at coarser intervals and plotted without additional smoothing. Lines and shaded regions show the across-run mean and sample standard deviation at the corresponding environment steps.

\subsection{Additional benchmark results}
\label{app:additional_results}

Tables~\ref{tab:sequence_classification} and~\ref{tab:popgym} expose complementary demands on memory. Fixed spectral models preserve long pixel sequences well but are weaker on RepeatFirst. Input-dependent gates improve RepeatFirst in RG-LRU, GateLoop, and Mamba-3, yet these implementations are weaker on permuted MNIST and RepeatPrevious. SPARC combines strong performance in both settings. Input-dependent gating alone is therefore insufficient for long-horizon recall unless persistent modes are protected from cumulative drift.

\begin{table}[H]
\centering
\caption{Partially observable memory tasks in POPGym.}
\label{tab:popgym}
\small
\tabstyle\setlength{\tabcolsep}{2.5pt}
\begin{tabular}{@{}p{.13\linewidth}C{.13\linewidth}*{6}{C{\dimexpr(.74\linewidth-14\tabcolsep)/6\relax}}@{}}
\toprule
\textbf{Method} & \textbf{Training} & \shortstack{\textbf{Auto}\\\textbf{encode}} & \shortstack{\textbf{Count}\\\textbf{Recall}} & \shortstack{\textbf{Repeat}\\\textbf{First}} & \shortstack{\textbf{Repeat}\\\textbf{Previous}} & \shortstack{\textbf{Noisy}\\\textbf{Pendulum}} & \shortstack{\textbf{Higher}\\\textbf{Lower}}\\
\midrule
RTU & RTRL & \tabmean{-0.443}{0.017} & \tabmean{-0.540}{0.011} & \tabmean{0.432}{0.070} & \tabmean{0.843}{0.087} & \tabmean{0.441}{0.115} & \tabmean{0.496}{0.002}\\
LRU & RTRL & \tabmean{\underline{-0.424}}{0.042} & \tabmean{-0.537}{0.342} & \tabmean{0.017}{0.301} & \tabmean{\underline{0.884}}{0.088} & \tabmean{0.428}{0.025} & \tabmean{0.497}{0.004}\\
RG-LRU & RTRL & \tabmean{-0.473}{0.010} & \tabmean{-0.582}{0.088} & \tabmean{0.564}{0.073} & \tabmean{-0.079}{0.061} & \tabmean{0.467}{0.014} & {\boldmath\bfseries\tabmean{0.504}{0.001}}\\
GateLoop & RTRL & \tabmean{-0.425}{0.010} & \tabmean{-0.642}{0.043} & \tabmean{0.708}{0.036} & \tabmean{0.199}{0.065} & {\boldmath\bfseries\tabmean{0.543}{0.026}} & \tabmean{0.498}{0.001}\\
Mamba-3 & TBPTT-16 & \tabmean{-0.497}{0.002} & \tabmean{-0.669}{0.088} & \tabmean{0.605}{0.034} & \tabmean{0.347}{0.248} & \tabmean{0.355}{0.114} & \tabmean{0.491}{0.003}\\
Mamba-3 & RTRL & \tabmean{-0.498}{0.001} & \tabmean{-0.816}{0.083} & \tabmean{0.199}{0.085} & \tabmean{-0.293}{0.046} & \tabmean{0.197}{0.015} & \tabmean{0.471}{0.020}\\
GRU & TBPTT-16 & \tabmean{-0.472}{0.011} & {\boldmath\bfseries\tabmean{-0.401}{0.101}} & {\boldmath\bfseries\tabmean{0.862}{0.052}} & \tabmean{0.881}{0.080} & \tabmean{0.327}{0.076} & \tabmean{0.496}{0.000}\\
\tabgroup
\textbf{SPARC} & RTRL & {\boldmath\bfseries\tabmean{-0.421}{0.012}} & \tabmean{\underline{-0.471}}{0.045} & \tabmean{\underline{0.817}}{0.045} & {\boldmath\bfseries\tabmean{0.906}{0.060}} & \tabmean{\underline{0.513}}{0.012} & \tabmean{\underline{0.499}}{0.002}\\
\bottomrule
\end{tabular}
\tabnote{Returns average the final $100$ rollouts within each run; cells show the across-run mean $\pm$ sample SD. Best means are bold; second-best means are underlined.}
\end{table}

\subsection{Long Range Arena}
\label{app:lra_results}

Models use the same residual backbone and full-sequence BPTT on Image, ListOps, Text, and Retrieval from LRA \citep{tay2020lra}. Image and ListOps use $20{,}000$ training updates; Text and Retrieval use $5000$. Checkpoints are selected on the validation split and evaluated on the test split.

\begin{table}[H]
\centering
\caption{Test accuracy (\%) on four LRA tasks.}
\label{tab:lra}
\small
\tabstyle
\begin{tabular}{@{}p{.16\linewidth}*{4}{C{\dimexpr(\linewidth-.16\linewidth-8\tabcolsep)/4\relax}}@{}}
\toprule
\textbf{Method} & \textbf{Image} & \textbf{ListOps} & \textbf{Text} & \textbf{Retrieval}\\
\midrule
RTU & $62.630$ & $37.200$ & $82.764$ & $\underline{66.904}$\\
LRU & $61.690$ & $38.600$ & $82.224$ & $66.158$\\
RG-LRU & $56.850$ & $39.050$ & $76.988$ & $65.631$\\
GateLoop & $56.810$ & $39.000$ & $78.436$ & $63.474$\\
Mamba-3 & $31.730$ & $37.950$ & $62.860$ & $56.965$\\
GRU & $\underline{66.250}$ & {\boldmath$45.300$} & {\boldmath$87.188$} & $65.590$\\
\tabgroup
\textbf{SPARC} & {\boldmath$68.340$} & $\underline{39.350}$ & $\underline{83.216}$ & {\boldmath$66.921$}\\
\bottomrule
\end{tabular}
\end{table}

In Table~\ref{tab:lra}, SPARC leads on Image and Retrieval and exceeds the spectral and selective baselines on all four LRA tasks. GRU is stronger on ListOps and Text, where its nonlinear state update may offer useful representational flexibility. The Image gain over GRU is $2.09$ percentage points, whereas Retrieval is close to RTU.

\subsection{Language modeling and downstream evaluation}
\label{app:language_modeling}

Both decoders have eight layers, width $2048$, a SwiGLU feed-forward width of $5632$, RMS normalization, and tied input/output embeddings. SPARC uses $1024$ complex modes per layer; RG-LRU has $2048$ real state coordinates and $16$ gate blocks. Training uses FineWeb-Edu's \texttt{sample-10BT} subset \citep{lozhkov2024finewebedu}, tokenized with the GPT-2 tokenizer (vocabulary $50{,}257$). Documents are assigned to disjoint training, validation, and test streams by a hash of their text; streams contain approximately 5B, 2M, and 10M prediction tokens. Sequences have length $2048$, with state reset at each window boundary and end-of-document tokens retained inside windows.

\begin{figure}[H]
\centering
\includegraphics[width=0.98\linewidth]{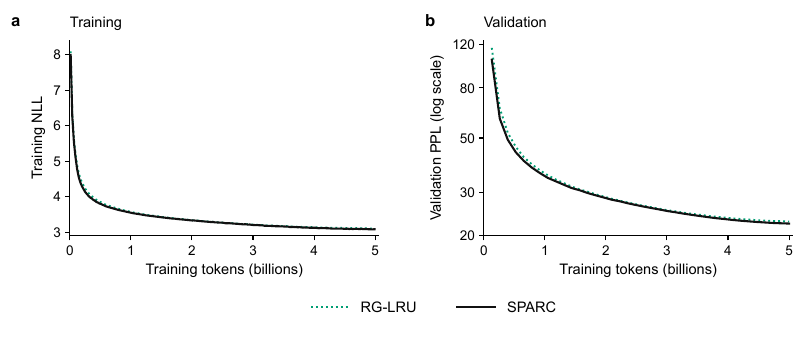}
\caption{Language-model convergence at $D=2048$: training NLL (100-update averages) and validation perplexity.}
\label{fig:language_model_curves}
\end{figure}

The effective batch is $128$ sequences ($262{,}144$ tokens). Each model trains for $19{,}073$ updates, totaling $4{,}999{,}872{,}512$ tokens. AdamW uses $\beta_1=0.9$, $\beta_2=0.95$, $\epsilon=10^{-8}$, gradient-norm clipping at $1$, and dense-weight decay $0.1$. Both models use a peak learning rate of $4\times10^{-4}$, with $2\%$ warmup and cosine decay to one tenth of the peak. SPARC's modal parameters use its peak rate and its controllers one quarter; RG-LRU's decay parameters use one quarter of its dense rate. SPARC initializes radii in $[0.9,0.999]$ and baseline frequencies below $0.003$ radians. Figure~\ref{fig:language_model_curves} shows convergence; Tables~\ref{tab:language_modeling} and~\ref{tab:zero_shot} report held-out and downstream performance. Downstream evaluation in Table~\ref{tab:zero_shot} tests the final pretrained checkpoints on standard zero-shot prompts without task-specific fine-tuning, scoring candidate completions by total conditional log-likelihood without length normalization.

\begin{table}[H]
\centering
\caption{Language modeling at $D=2048$. Lower is better.}
\label{tab:language_modeling}
\normalsize
\tabstyle
\renewcommand{\arraystretch}{1.3}
\begin{tabular*}{.80\linewidth}{@{\extracolsep{\fill}}lrr@{}}
\toprule
\textbf{Model} & \textbf{Test PPL} $\downarrow$ & \textbf{Test NLL} $\downarrow$\\
\midrule
\textbf{SPARC} & {\boldmath$22.136$} & {\boldmath$3.0972$}\\
RG-LRU & $22.588$ & $3.1174$\\
\bottomrule
\end{tabular*}
\end{table}

\begin{table}[H]
\centering
\caption{Zero-shot downstream accuracy (\%), without length normalization.}
\label{tab:zero_shot}
\small
\tabstyle
\setlength{\tabcolsep}{3pt}
\begin{tabular*}{\linewidth}{@{\extracolsep{\fill}}l*{8}{c}@{}}
\toprule
\textbf{Model} & \textbf{PIQA} & \textbf{HellaSwag} & \textbf{ARC-E} & \textbf{ARC-C} & \textbf{BoolQ} & \textbf{LAMBADA} & \textbf{RTE} & \textbf{WinoGrande}\\
\midrule
\textbf{SPARC} & {\boldmath$65.669$} & {\boldmath$30.592$} & {\boldmath$57.323$} & {\boldmath$22.696$} & {\boldmath$61.162$} & {\boldmath$7.588$} & {\boldmath$53.430$} & $50.671$\\
RG-LRU & $63.602$ & $29.974$ & $56.355$ & $22.611$ & $55.199$ & $7.122$ & $50.903$ & {\boldmath$51.776$}\\
\bottomrule
\end{tabular*}
\end{table}

\begin{table}[H]
\centering
\caption{Median language-model update time on two H800 GPUs.}
\label{tab:language_model_timing}
\small
\tabstyle
\begin{tabular}{@{}p{.50\linewidth}C{\dimexpr\linewidth-.50\linewidth-2\tabcolsep\relax}@{}}
\toprule
\textbf{Model} & \textbf{Median s/update}\\
\midrule
SPARC & {\boldmath$1.92$}\\
RG-LRU & $2.29$\\
\bottomrule
\end{tabular}
\end{table}

In Table~\ref{tab:language_model_timing}, medians cover $5980$ post-warmup updates per model, including data preparation, host-to-device transfer, forward and backward computation, distributed gradient synchronization, clipping, and AdamW. Validation and checkpoint I/O are excluded.

\section{Component Ablations}
\label{app:measured_ablations}

We compare retention-only and phase-only transitions, a static transition, removal of the phase clock, and a static phase clock. The interventions are defined in Table~\ref{tab:component_ablation_definitions}. State capacity, backbone, and task budgets are matched. The static-clock variant replaces $(1-\rho_{j,t})d_t$ by $(1-e^{-\nu_j})d_t$ while retaining adaptive retention and writing.

\begin{table}[H]
\centering
\caption{Classification accuracy (\%) in core component ablations.}
\label{tab:component_main}
\label{tab:ablation_mech_clock_none}
\small
\tabstyle
\begin{tabular}{@{}p{.22\linewidth}*{5}{C{\dimexpr(\linewidth-.22\linewidth-10\tabcolsep)/5\relax}}@{}}
\toprule
\textbf{Variant} & \textbf{FordA} & \textbf{SLC} & \textbf{UWave} & \textbf{pMNIST} & \textbf{sCIFAR-10}\\
\midrule
\textbf{SPARC} & \tabmean{\underline{94.66}}{0.16} & {\boldmath\bfseries\tabmean{97.14}{0.03}} & \tabmean{\underline{94.29}}{0.83} & \tabmean{96.67}{0.29} & {\boldmath\bfseries\tabmean{63.11}{1.26}}\\
\tabgroup
Retention only & \tabmean{94.62}{0.11} & \tabmean{\underline{97.01}}{0.03} & \tabmean{94.20}{0.85} & {\boldmath\bfseries\tabmean{96.76}{0.20}} & \tabmean{61.54}{1.06}\\
Phase only & \tabmean{\underline{94.66}}{0.37} & \tabmean{96.65}{0.43} & \tabmean{93.42}{1.15} & \tabmean{\underline{96.74}}{0.15} & \tabmean{61.42}{0.93}\\
Static transition & {\boldmath\bfseries\tabmean{94.92}{0.21}} & \tabmean{96.75}{0.32} & \tabmean{93.27}{1.21} & \tabmean{96.62}{0.27} & \tabmean{61.04}{1.45}\\
No phase clock & \tabmean{94.28}{0.05} & \tabmean{94.78}{1.08} & \tabmean{93.05}{1.08} & \tabmean{96.31}{0.20} & \tabmean{50.77}{3.38}\\
Static phase clock & \tabmean{94.36}{0.80} & \tabmean{96.86}{0.37} & {\boldmath\bfseries\tabmean{95.09}{0.51}} & \tabmean{96.73}{0.20} & \tabmean{\underline{61.94}}{0.66}\\
\bottomrule
\end{tabular}
\end{table}

In Table~\ref{tab:component_main}, removing the clock causes the largest loss on sequential CIFAR-10, while the static clock recovers most of it. This separates the value of attenuating phase corrections from the additional benefit of coupling them to current retention. The full model leads StarLightCurves and sequential CIFAR-10, but individual simplifications remain competitive elsewhere: static transitions favor FordA, a static clock favors UWave, and retention-only has the highest permuted-MNIST mean.

\begin{table}[H]
\centering
\caption{Core component ablations in continuous control.}
\label{tab:core_continuous_control}
\small
\tabstyle
\begin{tabular}{@{}p{.24\linewidth}*{4}{C{\dimexpr(\linewidth-.24\linewidth-8\tabcolsep)/4\relax}}@{}}
\toprule
\textbf{Variant} & \textbf{Ant-P} & \textbf{Walker-P} & \textbf{Hopper-P} & \textbf{Cheetah-P}\\
\midrule
\textbf{SPARC} & \tabmean{\underline{4665.31}}{457.85} & {\boldmath\bfseries\tabmean{1153.70}{322.29}} & {\boldmath\bfseries\tabmean{1378.90}{139.94}} & {\boldmath\bfseries\tabmean{2868.21}{24.12}}\\
\tabgroup
Retention only & {\boldmath\bfseries\tabmean{4743.69}{391.29}} & \tabmean{894.66}{61.30} & \tabmean{\underline{1322.08}}{105.28} & \tabmean{2481.20}{19.50}\\
Phase only & \tabmean{4210.84}{252.31} & \tabmean{715.22}{153.05} & \tabmean{1214.73}{123.17} & \tabmean{2552.05}{0.63}\\
Static transition & \tabmean{4522.38}{306.73} & \tabmean{893.84}{226.56} & \tabmean{1246.81}{72.92} & \tabmean{2584.69}{101.78}\\
No phase clock & \tabmean{4009.03}{539.58} & \tabmean{915.54}{260.21} & \tabmean{1213.07}{72.11} & \tabmean{2541.74}{69.14}\\
Static phase clock & \tabmean{4013.55}{258.66} & \tabmean{\underline{923.81}}{196.03} & \tabmean{1197.09}{189.23} & \tabmean{\underline{2618.26}}{8.33}\\
\bottomrule
\end{tabular}
\end{table}

In Table~\ref{tab:core_continuous_control}, joint retention and phase adaptation gives the strongest means on Walker-P, Hopper-P, and Cheetah-P; retention-only is strongest on Ant-P. The adaptive clock also exceeds the static clock on each task, unlike the mixed classification pattern. Policy learning in these partially observed environments therefore benefits from adjusting the phase scale with the current retention level, although cross-run variation remains substantial on several tasks.

\begin{table}[H]
\centering
\caption{Core component ablations in POPGym.}
\label{tab:core_popgym_clock}
\small
\tabstyle
\begin{tabular}{@{}p{.20\linewidth}*{6}{C{\dimexpr(\linewidth-.20\linewidth-12\tabcolsep)/6\relax}}@{}}
\toprule
\textbf{Variant} & \shortstack{\textbf{Auto}\\\textbf{encode}} & \shortstack{\textbf{Count}\\\textbf{Recall}} & \shortstack{\textbf{Repeat}\\\textbf{First}} & \shortstack{\textbf{Repeat}\\\textbf{Previous}} & \shortstack{\textbf{Noisy}\\\textbf{Pendulum}} & \shortstack{\textbf{Higher}\\\textbf{Lower}}\\
\midrule
\textbf{SPARC} & {\boldmath\bfseries\tabmean{-0.421}{0.012}} & \tabmean{-0.471}{0.045} & \tabmean{\underline{0.817}}{0.045} & \tabmean{0.906}{0.060} & {\boldmath\bfseries\tabmean{0.513}{0.012}} & \tabmean{0.499}{0.002}\\
\tabgroup
Retention only & \tabmean{-0.438}{0.008} & \tabmean{-0.465}{0.036} & \tabmean{0.563}{0.282} & \tabmean{0.888}{0.126} & \tabmean{0.458}{0.079} & \tabmean{0.503}{0.005}\\
Phase only & \tabmean{-0.433}{0.033} & \tabmean{-0.460}{0.024} & \tabmean{0.689}{0.093} & \tabmean{\underline{0.924}}{0.031} & \tabmean{0.409}{0.136} & \tabmean{\underline{0.505}}{0.001}\\
Static transition & \tabmean{-0.446}{0.017} & {\boldmath\bfseries\tabmean{-0.442}{0.059}} & \tabmean{0.446}{0.353} & {\boldmath\bfseries\tabmean{0.933}{0.021}} & \tabmean{\underline{0.497}}{0.033} & {\boldmath\bfseries\tabmean{0.508}{0.001}}\\
No phase clock & \tabmean{-0.443}{0.012} & \tabmean{-0.722}{0.123} & \tabmean{0.350}{0.255} & \tabmean{0.647}{0.208} & \tabmean{0.380}{0.108} & \tabmean{\underline{0.505}}{0.002}\\
Static phase clock & \tabmean{\underline{-0.432}}{0.031} & \tabmean{\underline{-0.456}}{0.120} & {\boldmath\bfseries\tabmean{0.866}{0.017}} & \tabmean{0.883}{0.101} & \tabmean{0.449}{0.074} & \tabmean{0.501}{0.002}\\
\bottomrule
\end{tabular}
\tabnote{Returns average the final 100 rollouts within each run.}
\end{table}

In Table~\ref{tab:core_popgym_clock}, RepeatFirst is strongest with a static clock among the interventions. Removing the clock is weaker than retaining either adaptive control alone, supporting attenuation when preserving an early observation through later inputs. Autoencode differences are small relative to cross-run variation, while a static transition remains competitive on several tasks.

\section{RTRL Resource Measurements}
\label{app:rtrl_resources}

\begin{table}[H]
\centering
\caption{RTRL resource cost for Ant-P: input width $D=64$ and $384$ real state coordinates ($H=192$ complex modes for SPARC).}
\label{tab:rtrl_resources}
\small
\tabstyle
\begin{tabular}{@{}p{.18\linewidth}*{4}{C{\dimexpr(\linewidth-.18\linewidth-8\tabcolsep)/4\relax}}@{}}
\toprule
\textbf{Method} & \shortstack{\textbf{Update time}\\\textbf{(s)}} & \shortstack{\textbf{GPU memory}\\\textbf{(MiB)}} & \shortstack{\textbf{Host trace}\\\textbf{(MiB)}} & \shortstack{\textbf{Running trace}\\\textbf{(MiB)}}\\
\midrule
SPARC & {\boldmath$3.80$} & {\boldmath$346$} & {\boldmath$12.33$} & {\boldmath$0.39$}\\
GRU & $198.17$ & $8532$ & $24264$ & $758.25$\\
Mamba-3 & $\underline{6.23}$ & $\underline{452}$ & $\underline{426.54}$ & $\underline{13.33}$\\
\bottomrule
\end{tabular}
\end{table}

In Table~\ref{tab:rtrl_resources}, host trace is the CPU cache of sensitivity traces for PPO updates, running trace is the sensitivity state at the current step, and GPU memory is the sampled device allocation.

SPARC's diagonal structure keeps the running trace at $0.39$ MiB, compared with $13.33$ MiB for Mamba-3 and $758.25$ MiB for dense GRU RTRL. The corresponding update times are $3.80$, $6.23$, and $198.17$ seconds. This illustrates the practical cost of carrying sensitivities through more densely coupled recurrent states.

\section{Control Sharing Ablation}
\label{app:unshared_controls}

We vary the number of control groups $G\in\{1,4,16,H\}$. Each group shares a retention--phase pair $(c_t,d_t)$, giving $2G$ dynamic scalars per layer; $G=1$ is global sharing and $G=H$ gives independent controls for each complex mode. State capacity, the baseline spectrum, write parameterization, backbone, and training budget are matched.

\begingroup
\footnotesize
\tabstyle\setlength{\tabcolsep}{2.5pt}
\begin{longtable}{@{}p{.27\linewidth}*{4}{C{\dimexpr(\linewidth-.27\linewidth-8\tabcolsep)/4\relax}}@{}}
\caption{Control grouping at matched state capacity and training budget.}\label{tab:ablation_unshared_controls}\\
\toprule
\textbf{Task} & \shortstack{\textbf{Shared}\\$G=1$} & \shortstack{\textbf{Grouped}\\$G=4$} & \shortstack{\textbf{Grouped}\\$G=16$} & \shortstack{\textbf{Unshared}\\$G=H$}\\
\midrule
\endfirsthead
\multicolumn{5}{@{}l}{\tablename~\thetable\ (continued)}\\[2pt]
\toprule
\textbf{Task} & \shortstack{\textbf{Shared}\\$G=1$} & \shortstack{\textbf{Grouped}\\$G=4$} & \shortstack{\textbf{Grouped}\\$G=16$} & \shortstack{\textbf{Unshared}\\$G=H$}\\
\midrule
\endhead
\midrule
\multicolumn{5}{r@{}}{Continued on next page}\\
\endfoot
\bottomrule
\endlastfoot
Autoencode & \tabmean{-0.421}{0.012} & \tabmean{-0.438}{0.017} & {\boldmath\bfseries\tabmean{-0.400}{0.019}} & \tabmean{\underline{-0.408}}{0.010}\\
CountRecall & \tabmean{\underline{-0.471}}{0.045} & \tabmean{-0.505}{0.085} & \tabmean{-0.537}{0.025} & {\boldmath\bfseries\tabmean{-0.416}{0.032}}\\
RepeatFirst & {\boldmath\bfseries\tabmean{0.817}{0.045}} & \tabmean{0.777}{0.220} & \tabmean{0.610}{0.434} & \tabmean{\underline{0.783}}{0.155}\\
RepeatPrevious & {\boldmath\bfseries\tabmean{0.906}{0.060}} & \tabmean{0.750}{0.058} & \tabmean{\underline{0.854}}{0.124} & \tabmean{0.727}{0.247}\\
Noisy Pendulum & {\boldmath\bfseries\tabmean{0.513}{0.012}} & \tabmean{\underline{0.461}}{0.043} & \tabmean{0.422}{0.009} & \tabmean{0.328}{0.108}\\
HigherLower & \tabmean{0.499}{0.002} & \tabmean{\underline{0.505}}{0.002} & \tabmean{0.504}{0.010} & {\boldmath\bfseries\tabmean{0.513}{0.005}}\\
\tabgroup
Ant-P & {\boldmath\bfseries\tabmean{4665.31}{457.85}} & \tabmean{3565.62}{530.00} & \tabmean{\underline{4212.76}}{336.23} & \tabmean{4023.15}{767.38}\\
Walker-P & {\boldmath\bfseries\tabmean{1153.70}{322.29}} & \tabmean{941.08}{232.11} & \tabmean{\underline{995.28}}{94.39} & \tabmean{943.07}{197.16}\\
Hopper-P & {\boldmath\bfseries\tabmean{1378.90}{139.94}} & \tabmean{\underline{1333.01}}{81.39} & \tabmean{1323.96}{203.64} & \tabmean{1082.24}{275.25}\\
Cheetah-P & \tabmean{\underline{2868.21}}{24.12} & \tabmean{2641.21}{44.56} & \tabmean{2489.62}{35.89} & {\boldmath\bfseries\tabmean{2888.18}{29.79}}\\
\tabgroup
FordA & {\boldmath\bfseries\tabmean{94.66}{0.16}} & \tabmean{\underline{94.55}}{0.32} & \tabmean{94.36}{0.70} & \tabmean{94.32}{0.64}\\
StarLightCurves & {\boldmath\bfseries\tabmean{97.14}{0.03}} & \tabmean{96.70}{0.64} & \tabmean{96.76}{0.22} & \tabmean{\underline{96.87}}{0.33}\\
UWave & \tabmean{94.29}{0.83} & \tabmean{94.41}{0.78} & \tabmean{\underline{94.62}}{0.60} & {\boldmath\bfseries\tabmean{94.92}{0.80}}\\
Permuted MNIST & \tabmean{\underline{96.67}}{0.29} & \tabmean{96.62}{0.11} & \tabmean{96.61}{0.13} & {\boldmath\bfseries\tabmean{96.75}{0.29}}\\
Seq. CIFAR-10 Gray & {\boldmath\bfseries\tabmean{63.11}{1.26}} & \tabmean{61.08}{0.85} & \tabmean{62.79}{1.57} & \tabmean{\underline{62.88}}{1.12}\\
\end{longtable}
\tabnote{Classification scores are test accuracies (\%); other tasks report returns. POPGym returns average the final 100 rollouts within each run. Higher is better. Best means are bold; second-best means are underlined.}
\endgroup

Table~\ref{tab:ablation_unshared_controls} covers all $15$ main tasks, with $H=192$ for continuous control and $H=64$ for the remaining tasks. Classification uses the complete training partition and reports test accuracy at the fixed $20{,}000$-update endpoint. PPO uses the same task-specific training budgets as the main experiments.

Shared control has higher means and smaller cross-run standard deviations than independent control on Ant-P and Hopper-P. These patterns are consistent with a regularizing effect of low-dimensional control in noisy online learning. Independent controls help Cheetah-P, showing that mode-specific adaptation can still be useful within continuous control. Each mode retains its own $\nu_j,\theta_j,s_j,v_j$ under sharing, preserving heterogeneous timescales and responses without a separate controller for every coordinate.

In classification, shared control has higher means on FordA, StarLightCurves, and sequential CIFAR-10; on sequential CIFAR-10 it exceeds independent controls by $0.23$ percentage points. Independent controls improve permuted MNIST and UWave by $0.08$ and $0.63$ percentage points, respectively. Shared control therefore retains competitive accuracy with a much smaller control pathway.

Each recurrent layer uses $2G(D+1)$ control-projection parameters. At $D=64$, $G=1,4,16$ use $130$, $520$, and $2080$ parameters, respectively. Independent controls use $8320$ parameters for $H=64$ or $24{,}960$ for $H=192$. Sharing reduces control-projection work from $O(HD)$ to $O(D)$ while retaining the same recurrent-state capacity.

\subsection{Cost of control grouping}
\label{app:control_cost_isolation}
\label{app:gpu_reference_checks}

All four settings use the same accelerated SPARC implementation: the two control heads are packed into one projection, followed by fused writing and the chunk-$32$ summary--prefix--replay scan and backward kernels. Modal transitions are reconstructed inside the scan. State capacity, the baseline spectrum, writing, and readout are matched. Outputs and gradients were checked against an independent complex-valued reference and the original kernels. All configurations were measured on one A800 GPU.

\begin{table}[H]
\centering
\caption{Task-scale control cost: $D=H=64$, $B=32$, $T=1024$, FP32.}
\label{tab:control_cost_task}
\small
\tabstyle
\begin{tabular}{@{}p{.28\linewidth}*{3}{C{\dimexpr(\linewidth-.28\linewidth-6\tabcolsep)/3\relax}}@{}}
\toprule
\textbf{Control setting} & \shortstack{\textbf{Control}\\\textbf{parameters}} & \shortstack{\textbf{Mixer F+B}\\\textbf{(ms)}} & \shortstack{\textbf{Peak allocated}\\\textbf{memory (MiB)}}\\
\midrule
Shared ($G=1$) & $130$ & {\boldmath$2.657$} & {\boldmath$137.478$}\\
Grouped ($G=4$) & $520$ & $\underline{2.664}$ & $\underline{140.106}$\\
Grouped ($G=16$) & $2080$ & $2.742$ & $150.618$\\
Unshared ($G=H$) & $8320$ & $2.913$ & $192.665$\\
\bottomrule
\end{tabular}
\end{table}

At task width, sharing chiefly reduces coefficient storage: peak allocated memory falls by $28.64\%$ relative to Unshared, while latency decreases by $8.79\%$ (Table~\ref{tab:control_cost_task}). The latency gap to $G=4$ is small, and grows as the number of independently projected controls increases. StarLightCurves accuracy is maintained at $97.14\%$, compared with $96.87\%$ for independent controls at this state capacity.

\begin{table}[H]
\centering
\caption{Wide-layer control cost: $D=2048$, $H=1024$, $B=4$, $T=2048$, BF16.}
\label{tab:control_cost_wide}
\small
\tabstyle
\begin{tabular}{@{}p{.28\linewidth}*{3}{C{\dimexpr(\linewidth-.28\linewidth-6\tabcolsep)/3\relax}}@{}}
\toprule
\textbf{Control setting} & \shortstack{\textbf{Control}\\\textbf{parameters}} & \shortstack{\textbf{Mixer F+B}\\\textbf{(ms)}} & \shortstack{\textbf{Peak allocated}\\\textbf{memory (MiB)}}\\
\midrule
Shared ($G=1$) & $4098$ & {\boldmath$2.704$} & {\boldmath$336.482$}\\
Grouped ($G=4$) & $16{,}392$ & $\underline{2.783}$ & $\underline{337.044}$\\
Grouped ($G=16$) & $65{,}568$ & $2.879$ & $339.294$\\
Unshared ($G=H$) & $4{,}196{,}352$ & $14.644$ & $534.317$\\
\bottomrule
\end{tabular}
\end{table}

At $D=2048$, the wider control projections make sharing more consequential (Table~\ref{tab:control_cost_wide}). Mixer forward--backward latency falls from $14.644$ to $2.704$ ms, a $5.41\times$ speedup, with $37.03\%$ lower peak allocated memory. With 8K tokens per batch, the FP32 control streams occupy $0.0625$ MiB for Shared and $64$ MiB for Unshared. Reconstructing modal transitions from the shared streams reduces both projection work and the volume of intermediate controls.

Relative to $G=4$ and $G=16$, Shared uses $75\%$ and $93.75\%$ fewer control parameters. At this width, its latency is lower by approximately $2.83\%$ and $6.08\%$, and memory is lower by $0.17\%$ and $0.83\%$, respectively. The default $G=1$ also avoids choosing a control-group hyperparameter.

\section{GPU Implementation Details}
\label{app:gpu_implementation_details}

Algorithm~\ref{alg:chunked_scan} divides time into chunks $\mathcal C_k=[s_k,e_k]$ and modes into $M$ tiles, indexed by $\ell$. Each tile reconstructs the affine maps $F_{t,\ell}$ from Eq.~\eqref{eq:sparc_recurrence}.

\begin{algorithm}[H]
\caption{Chunked scan with backward replay.}
\label{alg:chunked_scan}
\small
\begin{algorithmic}[1]

\Statex \textbf{Input:} shared controls, writes, $\mathbf{h}_0$, state cotangents $\mathbf{u}_{1:T}$, chunks $\{\mathcal C_k\}_{k=1}^{K}$
\Statex \textbf{Output:} $\mathbf{h}_{1:T},\,\nabla\boldsymbol\psi$

\Statex $\blacktriangleright$ \textbf{\textsc{Forward Pass}}
\State \textbf{parallel for} $(k,\ell)\in[K]\times[M]$:
       $\quad S_{k,\ell}\gets\operatorname{Compose}(F_{\mathcal C_k,\ell})$
\State \textbf{parallel for} $\ell\in[M]$:
       $\quad \widehat S_{1:K,\ell}\gets
       \operatorname{PrefixScan}(S_{1:K,\ell})$
\State \textbf{parallel for} $(k,\ell)\in[K]\times[M]$:
       $\quad
       \mathbf{h}_{\mathcal C_k,\ell}\gets
       \operatorname{Replay}
       (F_{\mathcal C_k,\ell},
       \widehat S_{k-1,\ell}(\mathbf{h}_{0,\ell}))$

\Statex $\blacktriangleright$ \textbf{\textsc{Backward Pass}}
\State \textbf{parallel for} $(k,\ell)\in[K]\times[M]$:
       $\quad
       \widetilde S_{k,\ell}\gets
       \operatorname{AdjointSummary}(F_{\mathcal C_k,\ell},\mathbf{u}_{\mathcal C_k,\ell})$
\State \textbf{parallel for} $\ell\in[M]$:
       $\quad
       \widetilde P_{1:K,\ell}\gets
       \operatorname{ReverseScan}(\widetilde S_{1:K,\ell})$
\State \textbf{parallel for} $(k,\ell)\in[K]\times[M]$:
       $\quad
       (\bar{\mathbf{h}}_{s_k-1,\ell},\Delta\boldsymbol\psi_{k,\ell})
       \gets
       \operatorname{AdjointReplay}
       (F_{\mathcal C_k,\ell},\mathbf{u}_{\mathcal C_k,\ell},\widetilde P_{k+1,\ell}(\mathbf{0}))$
\State $\nabla\boldsymbol\psi
       \gets
       \operatorname{Reduce}_{k,\ell}(\Delta\boldsymbol\psi_{k,\ell})$

\end{algorithmic}
\end{algorithm}

\paragraph{Coefficient generation and layout.}
The complex state is stored in separate real and imaginary arrays. The two controller projections are evaluated once per token in FP32, producing $B\times T$ arrays for $c_t$ and $d_t$. Content projections run in parallel over tokens and modes. Each modal tile broadcasts the controls and loads its spectral and write parameters. Content activation, retention-dependent normalization, modal gating, and real--imaginary write conversion are fused into one tiled operation with the layout required by the scan.

\paragraph{Chunk summaries and replay.}
A GPU program assigned to a time chunk and a modal tile reconstructs the transitions and composes the local updates using Eq.~\eqref{eq:affine_composition}. It writes one complex-affine summary per chunk. A prefix scan gives the state entering each chunk, and local replay produces the complete state sequence. Episode resets set the transition to zero at the reset position, as in Eq.~\eqref{eq:app_masked_scan}. In Algorithm~\ref{alg:chunked_scan}, $\widehat S_{0,\ell}$ and $\widetilde P_{K+1,\ell}$ are identity maps, $\bar{\mathbf{h}}$ denotes a state cotangent, and $\Delta\boldsymbol\psi_{k,\ell}$ collects local modal-parameter, write, and shared-control gradient contributions.

\paragraph{Backward computation and storage.}
Each chunk forms a reverse summary of its cotangent propagation. A reverse scan supplies boundary cotangents $\widetilde P_{k+1,\ell}(\mathbf{0})$ entering step $e_k$ from chunk $k+1$ (zero at $k=K$), and local reverse replay reconstructs transitions and accumulates gradients. Modal contributions are reduced before updating the two token-level control gradients. The scan retains the state outputs needed by BPTT and recomputes transition coefficients during summary and replay. The write backward operation recomputes content activations, the normalization ratio, and the gate from saved inputs and controls, then reduces parameter gradients over token tiles.

\paragraph{Precision and online execution.}
BF16 inputs and outputs are supported. Controller projections, spectral parameters, exponentials, trigonometric functions, complex-affine summaries, recurrent accumulation, and reverse adjoints use FP32. Conversion to output precision occurs at write-back boundaries. For an online step, one fused kernel evaluates the two controls, reconstructs the transition and write, updates the real and imaginary states, and emits the visible activation and next FP32 state.

\subsection{Efficiency benchmark protocol}
\label{app:efficiency_protocol}

Both implementations use matched tensor shapes, BF16 inputs and outputs, and FP32 recurrent accumulation. Mixer timings include control and coefficient generation, recurrent propagation, and input and parameter gradients. Memory measurements compare peak allocated GPU memory for the same training workloads.

The benchmark dimensions are mixer width $D$, batch size $B$, and sequence length $T$. Figure~\ref{fig:gpu_efficiency}(a) fixes $BT=8192$ and $D=2048$, using the three sequence-length and batch-size pairs below. Panel (b) fixes $T=2048$ and $D=2048$, varying batch size from $1$ to $32$. Panel (c) estimates training time on two H800 GPUs from median update latency.

For complete recurrent-mixer forward and backward computation, the fixed-token configurations are
\begin{equation}
 (B,T)\in\{(4,2048),(2,4096),(1,8192)\},
 \qquad BT=8192.
 \label{eq:efficiency_token_budget}
\end{equation}
Latency reductions divide by the RG-LRU time at each matched configuration. The plotted bar heights are normalized by RG-LRU's latency at $(B,T)=(4,2048)$. The batch sweep fixes $T=2048$ and $D=2048$; its latency ratio crosses one between batch sizes $2$ and $4$ and is approximately $0.6$ at batch sizes $8$ and above. Peak allocated-memory ratios are approximately $0.85$--$0.87$ for most batch sizes.

\subsection{Speed versus a sequential PyTorch implementation}
\label{app:naive_performance}

To quantify the computational benefit of the accelerated kernels, we compare the SPARC mixer with an eager PyTorch implementation of the same recurrence. The naive implementation computes the control and write terms, unbinds the time axis once, and propagates the state sequentially; PyTorch automatic differentiation supplies the backward pass. Identical parameters, input tensors, and output cotangents are used for each pair. Both paths accept BF16 inputs, return BF16 outputs, accumulate recurrent states in FP32, and compute input and parameter gradients. An NVIDIA RTX PRO 6000 GPU runs the measurements with PyTorch 2.8.0, CUDA 12.8, and Triton 3.4.0. After compilation and one warmup iteration, three timed iterations yield the reported medians. CUDA synchronization brackets each timed stage.

At a fixed budget of $8192$ tokens, longer sequences increase the cost of sequential execution substantially (Table~\ref{tab:naive_fixed_tokens}). For width $2048$, naive backward latency rises from $238.730$\,ms at $T=2048$ to $967.711$\,ms at $T=8192$, while the accelerated measurements range from $0.911$ to $1.289$\,ms. The resulting backward speedups span $194.81$--$1061.91\times$. Width $2560$ exhibits a similar trend: its backward speedup increases from $218.45\times$ to $888.98\times$ over the same sequence lengths. Across both widths, combined forward--backward speedups range from $137.13\times$ to $869.62\times$. The measured gains show the practical benefit of the accelerated execution paths for full-sequence differentiation.

\begin{table}[H]
\centering
\caption{Accelerated SPARC versus sequential PyTorch at $BT=8192$.}
\label{tab:naive_fixed_tokens}
\small
\setlength{\tabcolsep}{3pt}
\renewcommand{\arraystretch}{1.15}
\begin{tabular*}{\linewidth}{@{\extracolsep{\fill}}rrr rrr rrr@{}}
\toprule
 & & & \multicolumn{3}{c}{Backward} & \multicolumn{3}{c}{Forward + backward}\\
\cmidrule(lr){4-6}\cmidrule(l){7-9}
$B$ & $T$ & $D$ & Naive & Accel. & Speedup & Naive & Accel. & Speedup\\
\midrule
4 & 2048 & 2048 & 238.730 & 1.225 & $194.81\times$ & 358.822 & 2.617 & $137.13\times$\\
2 & 4096 & 2048 & 479.999 & 1.289 & $372.50\times$ & 721.780 & 2.510 & $287.54\times$\\
1 & 8192 & 2048 & 967.711 & 0.911 & $1061.91\times$ & 1469.467 & 1.690 & $869.62\times$\\
\midrule
4 & 2048 & 2560 & 236.191 & 1.081 & $218.45\times$ & 358.953 & 1.824 & $196.78\times$\\
2 & 4096 & 2560 & 489.609 & 1.105 & $442.89\times$ & 740.985 & 1.846 & $401.39\times$\\
1 & 8192 & 2560 & 969.447 & 1.091 & $888.98\times$ & 1500.917 & 1.847 & $812.84\times$\\
\bottomrule
\end{tabular*}
\tabnote{Times are medians in milliseconds. Speedups divide unrounded naive and accelerated medians for the same configuration.}
\end{table}

Increasing batch size at fixed $T=2048$ and $D=2048$ produces a different scaling pattern (Table~\ref{tab:naive_batch}). Naive backward latency remains within $228.853$--$265.053$\,ms; the accelerated path grows from $0.853$\,ms at $B=1$ to $5.614$\,ms at $B=32$. Backward acceleration therefore falls from $268.35\times$ to $47.21\times$ between these endpoints. Combined speedup reaches $230.69\times$ at $B=4$ and remains $45.68\times$ at $B=32$. Together, the two sweeps demonstrate substantial acceleration across the tested sequence lengths and batch sizes, with the largest measured gains occurring in long-sequence, small-batch configurations.

\begin{table}[H]
\centering
\caption{Batch scaling of accelerated SPARC and sequential PyTorch: $T=D=2048$.}
\label{tab:naive_batch}
\small
\tabstyle
\begin{tabular*}{\linewidth}{@{\extracolsep{\fill}}r rrr rrr@{}}
\toprule
 & \multicolumn{3}{c}{Backward} & \multicolumn{3}{c}{Forward + backward}\\
\cmidrule(lr){2-4}\cmidrule(l){5-7}
$B$ & Naive & Accel. & Speedup & Naive & Accel. & Speedup\\
\midrule
1 & 228.853 & 0.853 & $268.35\times$ & 344.970 & 1.900 & $181.56\times$\\
2 & 240.696 & 0.942 & $255.42\times$ & 361.919 & 2.026 & $178.67\times$\\
4 & 236.798 & 0.890 & $265.99\times$ & 355.549 & 1.541 & $230.69\times$\\
8 & 239.711 & 1.710 & $140.21\times$ & 365.954 & 2.548 & $143.62\times$\\
16 & 244.437 & 3.044 & $80.31\times$ & 366.240 & 4.570 & $80.15\times$\\
32 & 265.053 & 5.614 & $47.21\times$ & 390.188 & 8.542 & $45.68\times$\\
\bottomrule
\end{tabular*}
\tabnote{Times are in milliseconds. The fixed-token and batch sweeps are measured in separate runs; speedups divide unrounded medians.}
\end{table}

\subsection{Numerical agreement with the sequential reference}
\label{app:naive_correctness}

Numerical validation uses an independent PyTorch reference that explicitly constructs the complex transition and write terms, advances the state through a sequential loop, and obtains derivatives through automatic differentiation. With $(B,T,D)=(2,64,32)$ and FP32 inputs, the accelerated mixer uses the chunked scan backend (chunk size 32). Nonzero random controller and write-gate parameters exercise input-dependent retention, phase, and writing. Both implementations receive matching parameters, inputs, and output cotangents, allowing direct comparison of the forward output, input gradient, and recurrent parameter gradients.

For each tensor $z$, we report the maximum absolute difference $\|z_{\mathrm{accel}}-z_{\mathrm{ref}}\|_\infty$ and its normalized value:
\begin{equation}
 \frac{\|z_{\mathrm{accel}}-z_{\mathrm{ref}}\|_\infty}
      {\max(1,\|z_{\mathrm{ref}}\|_\infty)}.
 \label{eq:gpu_numerical_error}
\end{equation}
Table~\ref{tab:naive_correctness} reports both measures for the output and each active gradient block. The maximum forward error is $5.66\times10^{-7}$, and normalized gradient errors range from $1.27\times10^{-7}$ to $1.43\times10^{-6}$. These errors confirm close numerical agreement in single precision.

\begin{table}[H]
\centering
\caption{Numerical agreement with the sequential PyTorch reference.}
\label{tab:naive_correctness}
\small
\tabstyle
\renewcommand{\arraystretch}{1.75}
\begin{tabular*}{\linewidth}{@{\extracolsep{\fill}}lrr@{}}
\toprule
Quantity & Maximum absolute error & Normalized error\\
\midrule
Forward output & $5.66\times10^{-7}$ & $5.66\times10^{-7}$\\
Input gradient & $2.03\times10^{-6}$ & $5.28\times10^{-7}$\\
Log-decay parameter gradient & $4.77\times10^{-6}$ & $1.17\times10^{-6}$\\
Log-phase parameter gradient & $7.63\times10^{-5}$ & $6.58\times10^{-7}$\\
Phase-controller gradient & $6.68\times10^{-6}$ & $7.09\times10^{-7}$\\
Retention-controller gradient & $3.05\times10^{-5}$ & $1.43\times10^{-6}$\\
Log-write-gain gradient & $1.43\times10^{-6}$ & $3.32\times10^{-7}$\\
Write-phase-response gradient & $2.38\times10^{-7}$ & $2.38\times10^{-7}$\\
Write-retention-response gradient & $1.27\times10^{-7}$ & $1.27\times10^{-7}$\\
\bottomrule
\end{tabular*}
\end{table}

\end{document}